\PassOptionsToPackage{table,xcdraw}{xcolor}
\documentclass[sigconf]{acmart}
\renewcommand\footnotetextcopyrightpermission[1]{}

\usepackage{amsmath,epsfig}
\usepackage{algorithm}

\usepackage{amssymb}
\usepackage{nccmath}
\usepackage{algpseudocode}
\usepackage{xspace}
\usepackage{color}
\usepackage{setspace}
\usepackage[normalem]{ulem}
\usepackage{bm}
\useunder{\uline}{\ul}{}
\usepackage{enumitem}
\usepackage{rotating}
\usepackage{booktabs}
\usepackage{multirow}

\usepackage{graphicx}  
\usepackage{tikz}
\usetikzlibrary{fit,shapes,arrows}  
\usepackage{subcaption}  
\usepackage[table,xcdraw]{xcolor} 
\usepackage[capitalize]{cleveref} 
\crefname{section}{Sec.}{Secs.}
\Crefname{section}{Section}{Sections}
\Crefname{table}{Table}{Tables}
\crefname{table}{Tab.}{Tabs.}

\makeatletter
\DeclareRobustCommand\onedot{\futurelet\@let@token\@onedot}
\def\@onedot{\ifx\@let@token.\else.\null\fi\xspace}

\newcommand{\algoName}{\textsc{FMTL-Bench}\xspace}
\AtBeginDocument{%
  }

\setcopyright{acmlicensed}
\copyrightyear{2018}
\acmYear{2018}
\acmDOI{XXXXXXX.XXXXXXX}

\acmConference[Conference acronym 'XX]{Make sure to enter the correct
  conference title from your rights confirmation emai}{June 03--05,
  2018}{Woodstock, NY}

\begin{document}

\title{Federated Multi-Task Learning on Non-IID Data Silos: An Experimental Study}

\author{Yuwen Yang}
\email{youngfish@sjtu.edu.cn}
\orcid{0000-0002-5405-5096}
\affiliation{%
  \institution{Shanghai Jiao Tong University}  
  \city{Shanghai}
  \country{China}
}

\author{Yuxiang Lu}
\email{luyuxiang_2018@sjtu.edu.cn}
\orcid{0009-0002-6344-3880}
\affiliation{%
  \institution{Shanghai Jiao Tong University}  
  \city{Shanghai}
  \country{China}
}

\author{Suizhi Huang}
\email{huangsuizhi@sjtu.edu.cn}
\orcid{0000-0003-0172-6711}
\affiliation{%
  \institution{Shanghai Jiao Tong University}  
  \city{Shanghai}
  \country{China}
}

\author{Shalayiding Sirejiding}
\email{salaydin@sjtu.edu.cn}
\orcid{0000-0003-1255-4994}
\affiliation{%
  \institution{Shanghai Jiao Tong University}  
  \city{Shanghai}
  \country{China}
}

\author{Hongtao Lu}
\email{htlu@sjtu.edu.cn}
\orcid{0000-0003-2300-3039}
\affiliation{%
  \institution{Shanghai Jiao Tong University}  
  \city{Shanghai}
  \country{China}
}

\author{Yue Ding}
\email{dingyue@sjtu.edu.cn}
\orcid{0000-0002-2911-1244}
\affiliation{%
  \institution{Shanghai Jiao Tong University}  
  \city{Shanghai}
  \country{China}
}

\renewcommand{\shortauthors}{Yuwen Yang et al.}

\begin{abstract}  
The innovative Federated Multi-Task Learning (FMTL) approach consolidates the benefits of Federated Learning (FL) and Multi-Task Learning (MTL), enabling collaborative model training on multi-task learning datasets. However, a comprehensive evaluation method, integrating the unique features of both FL and MTL, is currently absent in the field. This paper fills this void by introducing a novel framework, \algoName , for systematic evaluation of the FMTL paradigm. This benchmark covers various aspects at the data, model, and optimization algorithm levels, and comprises seven sets of comparative experiments, encapsulating a wide array of non-independent and identically distributed (Non-IID) data partitioning scenarios. We propose a systematic process for comparing baselines of diverse indicators and conduct a case study on communication expenditure, time, and energy consumption. Through our exhaustive experiments, we aim to provide valuable insights into the strengths and limitations of existing baseline methods, contributing to the ongoing discourse on optimal FMTL application in practical scenarios. The source code can be found on \href{https://github.com/youngfish42/FMTL-Benchmark}{https://github.com/youngfish42/FMTL-Benchmark}.
\end{abstract}



\maketitle
\begin{figure}[tbp]
    \centering
    \includegraphics[width=1\linewidth]{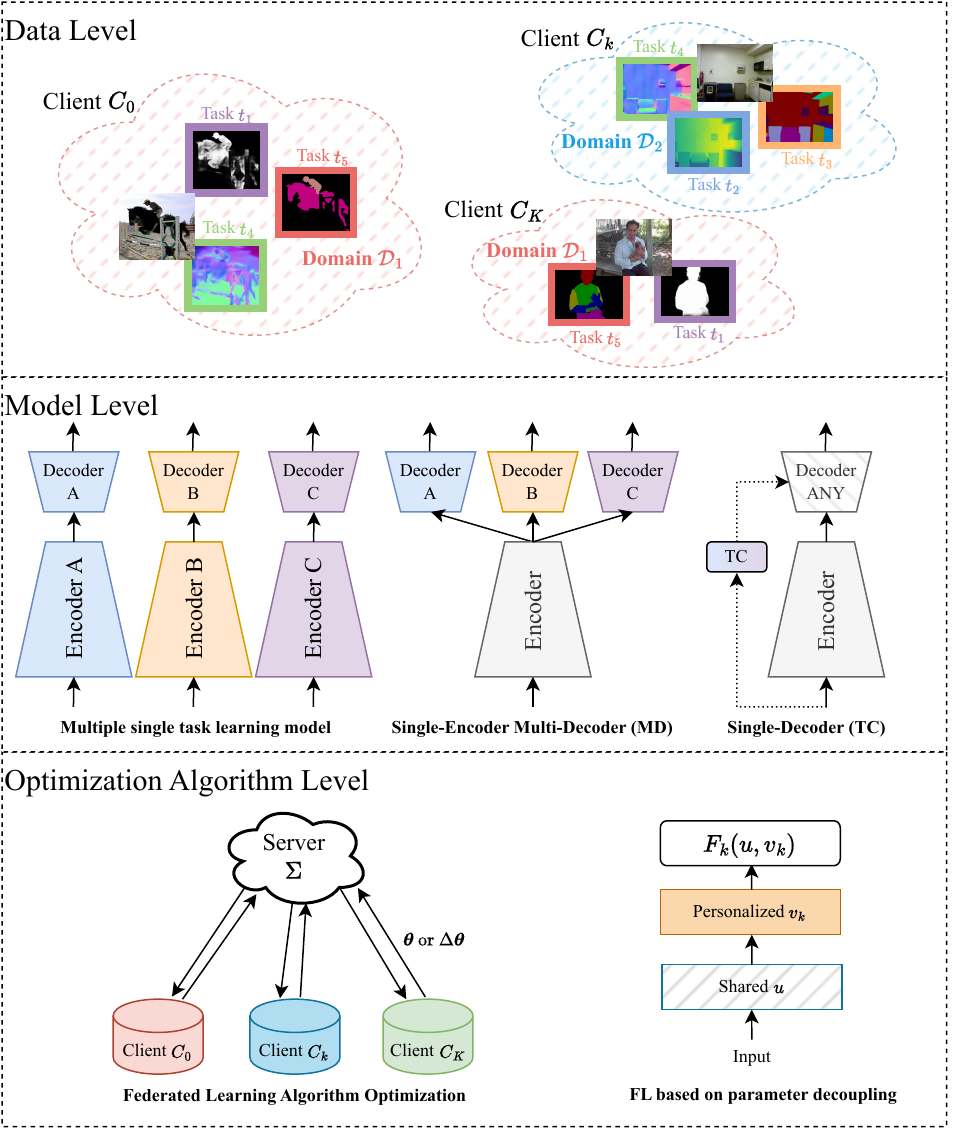}
    \caption{Design of Comparative Experimental Scenarios in \algoName. Refer to (\cref{sec: comp. exp}) for detailed information.}
    \begin{flushleft}
    \small
    \textbf{Data Level}: We design seven sets of experiments to cover main data partitioning scenarios in FMTL. These scenarios consider the different numbers and types of MTL tasks from various domains that a client may train. For more details, refer to~\cref{fig: comp. exp overview}. \textbf{Model Level}: We examine numerous single-task learning models and MTL models, the latter based on either multi-decoder (MD) or single-decoder architectures contingent on task conditions (TC). Experiments are conducted using network backbones of different sizes. The shaded sections in the figure represent task-agnostic parameters. \textbf{Optimization Algorithm Level}: We discuss nine baseline algorithms encompassing local training, FL, MTL, and FMTL algorithms. These algorithms leverage optimization based on either model parameters or accumulated gradients. Some baselines employ a parameter decoupling strategy and use model encoder as feature extractor during the FL process.
    \end{flushleft}
    \vspace{-10pt}
    \label{fig:overview-FMTL bench}
\end{figure}

\section{Introduction}
Federated Multi-Task Learning (FMTL)~\cite{MAS,THFMTL}, a burgeoning machine learning paradigm, facilitates collaborative model training on multi-task learning datasets of diverse sample sizes, domains, and task types while ensuring data locality. It marries the advantages of Federated Learning (FL)~\cite{1stfed,FedAvg} and Multi-Task Learning (MTL)~\cite{mtl1}, both extensively employed in sectors like medical imaging~\cite{hanBreakingMedicalData2020,kaissisEndtoendPrivacyPreserving2021,kaissisSecurePrivacypreservingFederated2020,flbrain,covid,qi_differentially_2023,MedPerf,feng_robustly_2024,jin_predicting_2021,eyuboglu_multi-task_2021}, healthcare~\cite{warnat-herresthalSwarmLearningDecentralized2021,zhangShiftingMachineLearning2022,covid2,wu_wearable_2023}, and personalized recommendations~\cite{wuCommunicationefficientFederatedLearning2022,wuFederatedGraphNeural2022,kalra_decentralized_2023}. FMTL enables a single model to learn multiple tasks in a privacy-preserving, distributed machine learning environment, thereby inheriting and amplifying the challenges of both FL~\cite{fedsurvey,fedsurvey2} and MTL~\cite{mtlsurvey, mtl2, mtl3}.

Early research~\cite{mtfl1,mtfl2,spreadgnn,FedMTL,fedem} predominantly adopted MTL’s optimization strategy for personalized federated learning (PFL)~\cite{pfedsurvey} contexts. In non-independent and identically distributed (Non-IID or NIID) scenarios, each client’s personalized optimization objective function was treated as an individual task, with MTL optimization methods managing task heterogeneity arising from diverse client data. Classic machine learning tasks like image classification were commonly employed. A handful of studies~\cite{FeSTA,MaT-FL,FedBone} have started probing more complex scenarios, aiming to allow clients to concurrently train different task types via FL. In some industrial contexts, such as autonomous driving~\cite{janai2020computer}, a single model must learn multiple distinct dense prediction tasks~\cite{deeplab, dpt, pvt} (e.g., semantic segmentation, depth estimation, and surface normal estimation) simultaneously. The final scenario~\cite{MAS,THFMTL} involves learning an MTL model within a single client, which is the focus of this article.

However, the current research landscape of the FMTL paradigm is still in its infancy and has not fully integrated the characteristics of both FL and MTL to establish suitable scenarios and evaluation methods. The currently employed FL and MTL optimization algorithms and models are relatively simple, and there is a scarcity of experimental research to systematically comprehend FMTL task scenarios and baselines. Given  growing interest in this technology, we introduce the \textsc{FMTL-Bench} to systematically evaluate the FMTL paradigm. Drawing from previous work on FL and MTL evaluation benchmarks~\cite{NIIDbench,MedPerf,MAS,carbonFootprint2023,mtlsurvey,pieri2024handling,hu2022oarf}, we integrated the strengths of these works to establish a comprehensive benchmark in the FMTL field, addressing data, model, and optimization algorithm levels for the first time.
\textbf{Contributions} of our paper include:
\begin{itemize}[leftmargin=*,noitemsep,topsep=0pt]
    \item We meticulously consider the data, model, and optimization algorithm to design seven sets of comparative experiments.
    \item We amalgamate the characteristics of the two fields of multi-task learning and federated learning, conduct a case study, and extensively utilize a variety of evaluation methods to assess the performance of each baseline.
    \item We glean insights from comparative experiments and case analyses, and provide application suggestions for FMTL scenarios.
\end{itemize}
 
\section{Methodology Overview} 
Suppose we have $K$ clients in total, Federated Learning (FL)~\cite{FedAvg} is an optimization process that aims to minimize a global objective function. This function is defined by model parameters $\theta_k$, local objective function $\mathcal{L}_{k}$, and client weights $p_k$. In the Federated Multi-Task Learning (FMTL) scenario, each client manages a set of tasks $\mathcal{T}_k$ associated with a local dataset. The global function seeks to optimize personalized models for each client as expressed in Equation~\ref{eq:fmtl}:  
\begin{equation}        
\label{eq:fmtl}
\begin{aligned}
    & \min_{\{\theta_k\}} \sum_{k \in [K]} p_k \mathcal{L}_{k}(\theta_k), \ \text{where\;}
    \\  &\mathcal{L}_{k}(\theta_k) = \frac{1}{\sum_{t\in \mathcal{T}_k} q_{k,t}} \sum_{t\in \mathcal{T}_k} q_{k,t} \ell_{k,t}(\theta_k).
\end{aligned}
\end{equation}   
  
In this equation, $\mathcal{L}_{k}$ is the local objective function. Weight $p_k$ of client $C_k$ in aggregation defaults to $\frac{1}{K}$, where $K$ is the total number of clients.  Each task loss function $\ell_{k,t}(\theta_k)$ is computed over the local dataset $\mathbb{D}_k$ of client $C_k$ for task $t$ in $\mathcal{T}_k$. The weights $q_{k,t}$ for each task $t$ default to $\frac{1}{|\mathcal{T}_k|}$, where $|\mathcal{T}_k|$ is the number of local tasks.

\textbf{Multi-Task Learning (MTL) model level architecture design.} To reduce the number of model parameters in MTL, a ``single-encoder multi-decoder'' (MD) architecture~\cite{mtlsurvey} is often employed. 
\begin{equation}\label{eq: MD}
\mathbf{y}_n=F_{k}^{\texttt{MD}}(\mathbf{x}_n; \theta_k^E, \sum_{t\in \mathcal{T}_k}\theta_{k,t}^D).
\end{equation}

Here, $\mathbf{x}_n$ denotes the $n$-th input data, $\mathbf{y}_n$ is the corresponding output, $\theta_{k}^E$ denotes the encoder parameters of the $k$-th client model, and $\theta_{k,t}^D$ represents the decoder parameters for task $t$. 

To further reduce the number of parameters, a ``single-decoder based on task conditions'' (TC) architecture~\cite{astmt, tsn} is utilized.   
\begin{equation}\label{eq: TC}
  \mathbf{y}_{n,t}=F_{k}^{\texttt{TC}}(\mathbf{x}_n; \theta_k^E,\theta_{k}^D,\theta_{k,t}^T),\quad \forall t\in \mathcal{T}_k.
\end{equation}

In this equation, $\theta_k^E$ and $\theta_k^D$ are encoder and decoder parameters shared among all tasks $\mathcal{T}_k$, and $\theta_{k,t}^T$ are the task-specific parameters for task $t$ used in the conditioning strategy. In addition to architecture design, MTL also requires a lot of optimization work from the perspective of parameters and gradients~\cite{mtlsurvey}.
  
\textbf{Federated learning (FL) algorithm optimization level.} Parameter decoupling strategies are also frequently used in FL and FMTL scenarios to reduce communication expenses, manage with model task heterogeneity, and improve optimization performance. These strategies divide the model parameters into shared parameters $u$ and personalized parameters $v_k$ for client $k$.   
\begin{equation}      
\label{eq:prameter decouple}
   \min_{u,\{v_k\}} \sum_{k \in [K]} p_k \mathcal{L}_{k}(u,v_k).
\end{equation}

In FL scenarios, sharing feature extractors between clients~\cite{fedrep} is common, while in FMTL scenarios, the encoder serves as a default feature extractor~\cite{FedBone,MaT-FL}. In addition to parameter decoupling strategy, we also verified the effects of various FL, PFL, MTL, and FMTL algorithms in subsequent experiments (see~\cref{sec:opt algo level}). We have introduced the core methodology, and due to space limitations, we will not discuss in detail the nine algorithms used in this article.

\begin{figure*}[tbp]  
    \begin{minipage}[c]{0.43\linewidth}  
        \centering 
        \resizebox{1\textwidth}{!}{
        \begin{tikzpicture}[transform shape, node distance = 2.6cm, auto, font=\footnotesize]  
        \tikzstyle{blockHighlighted} = [rectangle, draw=orange!60, fill=orange!5,   
        text width=6em, text centered, rounded corners, minimum height=3em] 
        \tikzstyle{SD block} = [rectangle, draw=gray!60, fill=gray!5,   
        text width=6em, text centered, rounded corners, minimum height=3em] 
        \tikzstyle{CD block} = [rectangle, draw=blue!60, fill=blue!5,   
        text width=6em, text centered, rounded corners, minimum height=3em]  
      
        \node [blockHighlighted] (IID-1) {IID-1 SDMT \\ (see \Cref{tab: Comp Exp IID-1 and NIID-3})}; 
        \node [SD block, below of=IID-1, yshift=0.4cm] (NIID-2 SDST) {NIID-2 SDST \\ (see \Cref{tab: Comp Exp NIID-2 SDST})};  
        \node [SD block, left of=IID-1, yshift=-1.1cm] (NIID-3 SDHT) {NIID-3 SDHT \\ (see \Cref{tab: Comp Exp IID-1 and NIID-3})};
        \node [SD block, right of=IID-1,xshift=.8cm] (NIID-4 UBSDMT) {NIID-4 UBSDMT \\ (see \Cref{tab: Comp Exp NIID-4 UBSDMT})};  
        \node [SD block, right of=NIID-2 SDST,xshift=.8cm] (NIID-5 UBSDST) {NIID-5 UBSDST \\ (see \Cref{tab: Comp Exp NIID-5 UBSDST})};  
        \node [CD block, above of=NIID-4 UBSDMT, yshift=-1.5cm] (NIID-6 UBCDMT) {NIID-6 UBCDMT \\ (see \Cref{tab: Comp Exp NIID-6 UBCDMT})}; 
        \node [CD block, below of=NIID-5 UBSDST, yshift= 1.5cm] (NIID-7 UBCDST) {NIID-7 UBCDST\\ (see \Cref{tab: Comp Exp NIID-7 UBCDST})}; 
        \draw [->] (IID-1) -- (NIID-2 SDST) node[midway, sloped] {single task};    
        \draw [->] (IID-1) -- (NIID-3 SDHT) node[midway, above, sloped] {merge};    
        \draw [->] (IID-1) -- (NIID-4 UBSDMT) node[midway, below] {unbalanced};    
        \draw [->] (NIID-2 SDST) -- (NIID-5 UBSDST) node[midway, above] {unbalanced};  
        \draw [->] (NIID-2 SDST) -- (NIID-3 SDHT) node[midway, below, sloped] {merge};   
        \draw [->] (IID-1) -- (NIID-6 UBCDMT) node[midway, above, sloped] {cross domain};    
        \draw [->] (NIID-2 SDST) -- (NIID-7 UBCDST) node[midway, below, sloped] {cross domain};

        \end{tikzpicture}
        }
    \end{minipage}  
    \hfill  
    \begin{minipage}[c]{0.56\linewidth}  
    \centering  
    \includegraphics[width=1\linewidth]{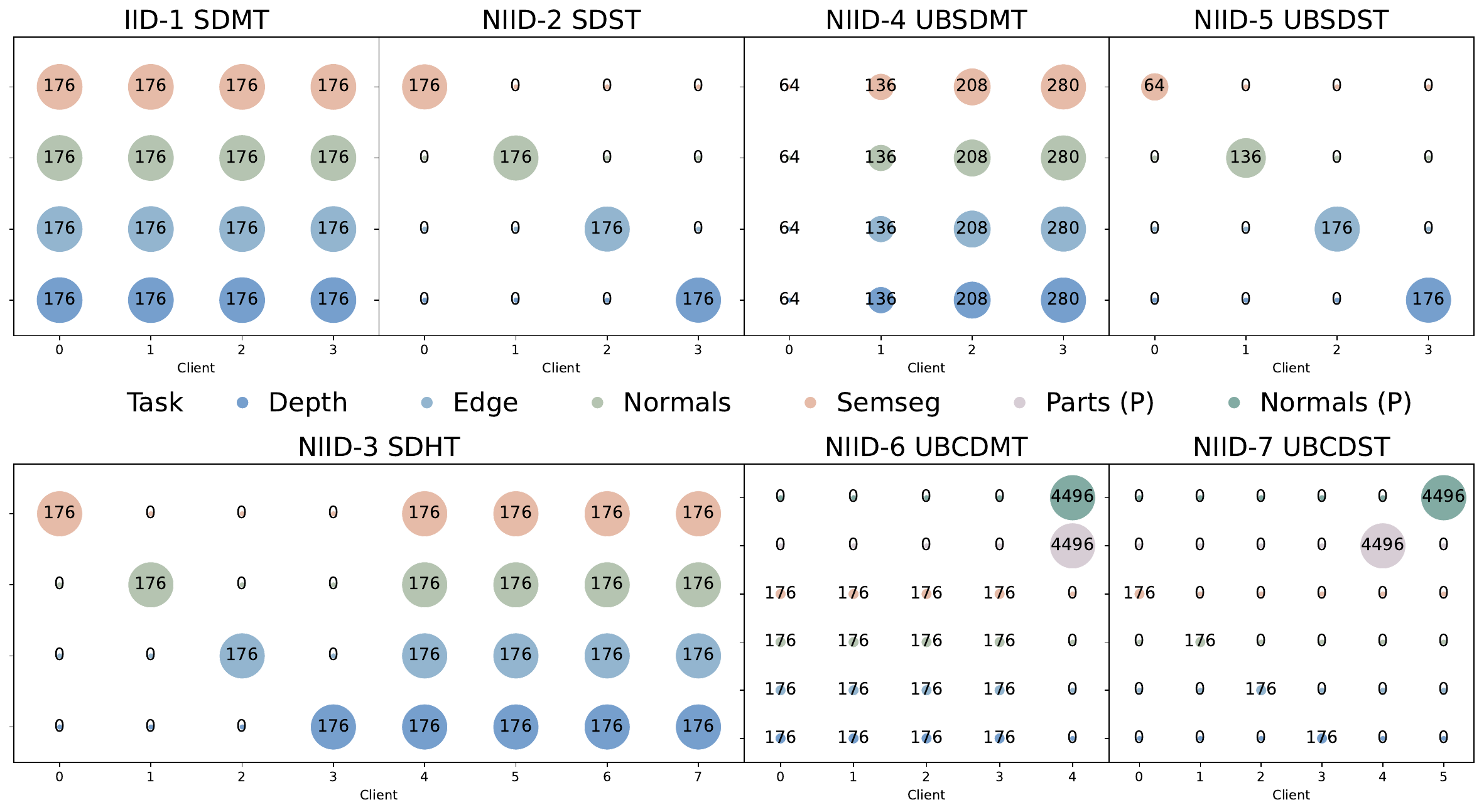}  
    \end{minipage}
    \vspace{-10pt}
    \caption{Data Level: Relationship diagram and Data Distribution of Comparative Experiments in Federated Multi-Task Learning.} 
    \begin{flushleft}  
    \small This figure presents two levels of detail regarding our comparative experiments in federated multi-task learning.  `MT' represents Multi-Task, `ST' stands for Single-Task, and `HT' denotes Hybrid-Task. `SD' signifies a single domain, while `CD' refers to cross-domain. `UB' is an abbreviation for unbalanced quantity. \textbf{Left-hand side}: an overview of the relationships between seven groups of comparative experiments is provided. This relationship diagram encapsulates the main scenarios of independent and identically distributed (IID) and non-independent and identically distributed (NIID) federated multi-task learning.  
    \textbf{Right-hand side}: a visualization of the training data distribution for these seven sets of comparative experiments is displayed. Each subfigure corresponds to a different comparative experiment. The horizontal axis denotes the client ID, while the vertical axis, differentiated by color, represents various types of tasks. Task types include depth estimation (`Depth'), edge detection (`Edge'), surface normal estimation (`Normals'), semantic segmentation (`SemSeg'), and human parts segmentation (`Parts'). Dataset comes from NYUD-v2 by default, and (P) represents using the PASCAL-Context. The relative size of the scatter points signifies the number of task samples, and the center of each point corresponds to the specific number of samples. For a detailed view, please zoom in.  
    \end{flushleft}
    \label{fig: comp. exp overview}
\end{figure*}  

\section{Experiment Evaluation}
This research aims to establish a comprehensive benchmark, referred to as \algoName, in the field of federated multi-task learning. We have designed a robust experimental setup with three key components: comparative experiment, case study, and suggestion.

\subsection{Experimental Setup}  
Our experiments are conducted using the PyTorch framework~\cite{pytorch}. The hardware setup comprises a server equipped with eight NVIDIA RTX2080Ti GPUs, while memory-intensive experiments are executed on two NVIDIA RTX4080 GPUs. The design of our experimental configurations is guided by previous studies~\cite{astmt, rcm, invpt, MaT-FL, FedBone, MAS}.    
  
\noindent\textbf{Optimization.}    
Model optimization is achieved using the AdamW optimizer~\cite{adamw}, with an initial learning rate and weight decay rate of 1e-4. The batch size is set to 8. A cosine decay learning rate scheduler~\cite{sgdr} is employed with a warm-up phase of 5 rounds. Loss functions are chosen based on the task; cross-entropy loss for semantic segmentation and human parts segmentation, and $\mathcal{L}_1$ loss for surface normal estimation and depth estimation. Weighted binary cross-entropy loss for edge detection and the weights for positive and negative pixels are set to 0.8 and 0.2 for the NYUD-v2 dataset, and 0.95 and 0.05 for the PASCAL-Context dataset. 

We adhere to the default settings for FL, PFL, MTL, and FMTL baselines (see~\cref{sec:opt algo level}) as recommended in the original papers or source codes. The number of local training epochs is set to 4 for the NYUD-v2 dataset and 1 for the PASCAL-Context dataset, based on dataset size. Maximum number of communication rounds is capped at 100. Source code will be made available to reproduce the results.

\noindent\textbf{Datasets.}
Our experiments utilize the PASCAL-Context~\cite{pascal} and NYUD-v2~\cite{nyud} datasets, both of which are widely recognized in FMTL research~\cite{MaT-FL, FedBone}.    
For single-domain experiments, we employ the NYUD-v2 dataset, which includes 795 training and 654 testing images of indoor scenes. The dataset provides labels for depth estimation (`Depth'), edge detection (`Edge'), surface normal estimation (`Normals'), and semantic segmentation (`SemSeg') tasks. In cross-domain experiments, we introduce the PASCAL-Context dataset, from which we obtain the same type of normal task (`Normals') as NYUD-v2 and a different type of human parts segmentation (`Parts'). 
The PASCAL-Context dataset contains 4998 training images and 5105 testing images for edge detection, semantic segmentation, human parts segmentation, surface normal estimation, and saliency detection tasks. 
The original training set is randomly divided into each client according to the specified method. The local dataset obtained by the client is re-divided into a local training set and a test set at a ratio of 9:1. Global evaluation G-FL and local evaluation P-FL~\cite{fedrod,shi2023prior} are conducted on the original test set and each client's dataset, respectively. Following~\cite{astmt, rcm, invpt}, we employ diverse data augmentation techniques  such as random scaling, cropping, horizontal flipping, and color jittering to augment training dataset. Both training and evaluation stages incorporate image normalization.

\subsection{Comparative Experiments}\label{sec: comp. exp}
Our comparative experiments comprehensively consider data, model, and optimization algorithms levels, among others.

\subsubsection{Data Level}\label{sec:data level}
We begin with the IID-1 SDMT (Single-Domain Multi-Task) experiment, an independent and identically distributed (IID) setup involving four clients, each possessing a non-overlapping dataset from the same domain for identical tasks. Motivated by the pathological partition scenario in federated learning~\cite{FedAvg}, we devised the NIID-2 SDST (Single-Domain Single-Task) scenario. In this setup, each client is responsible for a distinct task, representing an extreme case of federated multi-task learning. To examine the influence of the number of tasks within a client on the outcomes, we amalgamated the clients from the IID-1 and NIID-2 scenarios, leading to the NIID-3 SDHT (Single-Domain Hybrid-Task) scenario.
  
To investigate the impact of imbalanced data volume, we established the NIID-4 UBSDMT (Unbalanced Single-Domain Multi-Task) and NIID-5 UBSDST (Unbalanced Single-Domain Single-Task) scenarios, derived from the IID-1 and NIID-2 scenarios, respectively. For a comprehensive understanding of federated multi-task learning in cross-domain situations, we introduced the NIID-6 UBCDMT (Unbalanced Cross-Domain Multi-Task) and NIID-7 UBCDST (Unbalanced Cross-Domain Single-Task) scenarios. These scenarios involve clients from diverse fields, utilizing cross-domain data with unbalanced sample sizes, and exhibiting heterogeneous tasks and models. Refer to \cref{fig: comp. exp overview} for relationship diagram and visualization of training data distribution across comparative experiment groups.

\begin{table}[tbp]
\setlength{\tabcolsep}{2pt} 
\caption{Enumeration of multi-task learning model parameters and computational complexity (FLOPs) for clients in IID-1 Single-Domain Multi-Task (SDMT) scenario from~\cref{tab: Comp Exp IID-1 and NIID-3}.}
\small
\vspace{-10pt}
\resizebox{\linewidth}{!}{
\begin{tabular}{@{}l|c|ccccc|cc@{}}
\bottomrule
\multirow{2}{*}{Ar} & \multirow{2}{*}{BN} & \multicolumn{5}{c|}{Parameters (M)}                   & \multicolumn{2}{c}{FLOPs (G)} \\
                    &                           & Encoder & Decoder & TC module & Total & Encoder/Total & Per task       & Total        \\ \hline
MD                  & resnet                 & 11.18   & 26.21   & /      & 37.39 & 29.89\%       & -              & 272.14       \\
TC                  & resnet                 & 11.18   & 7.09    & 0.12      & 18.39 & 60.79\%       & 75.47          & 301.88       \\\hline
MD                  & swin-t                    & 27.52   & 52.28   & /     & 79.80 & 34.49\%       & -              & 211.13       \\
TC                  & swin-t                    & 27.52   & 13.76   & 0.12      & 41.40 & 66.48\%       & 71.01          & 284.04       \\ \toprule
\end{tabular}
}\label{tab:parameters and flops}
\end{table}

\subsubsection{Model Level}\label{sec:model level}  
In alignment with the conventional practice of implementing an ``encoder-decoder'' model architecture in multi-task learning, our experimental design subscribes to this paradigm. We bifurcate the client model into the encoder and decoder. 

In multi-task learning, the network typically shares a single encoder across multiple tasks. For this shared encoder, we employ a lightweight structure grounded on the pre-trained ResNet-18~\cite{resnet}. Additionally, we include pre-trained Swin-T~\cite{swin} in our experiments for comparative analysis. This backbone network is paired with a Fully Convolutional Network and the task-specific header.

The decoder component, on the other hand, can vary based on the architecture~\cite{mtlsurvey}. While the conventional ``multi-decoder'' (MD) architecture as~\cref{eq: MD} assigns a separate decoder to each task, we introduce an innovative ``single-decoder based on task conditions'' (TC) architecture~\cite{astmt, tsn} as~\cref{eq: TC}. This novel approach, applied in the context of FMTL, employs a single decoder that adjusts according to the specific conditions of each task.

\subsubsection{Optimization Algorithm Level}\label{sec:opt algo level}  
Our experimental setup deploys a suite of nine optimization algorithms, covering gradient- and parameter-based strategies for MTL optimization, along with personalization and parameter decoupling strategies for FL. 
The baseline algorithms are categorized as follows: Firstly, the \textbf{Local} method, which solely operates on local dataset training, refraining from involvement in the federated learning process. Secondly, the \textbf{Federated Learning Method}, exemplified by FedAvg~\cite{FedAvg}. Thirdly, the category of \textbf{Personalized Federated Learning Algorithms}, encompassing FedProx~\cite{fedprox}, FedAMP~\cite{fedamp}, and FedRep~\cite{fedrep}. Fourthly, \textbf{Multi-Task Learning Algorithms}, which includes PCGrad~\cite{PCGrad} and CAGrad~\cite{CAGrad}, both underpinned by gradient optimization techniques. Finally, the \textbf{Federated Multi-Task Learning Algorithms}, comprising the FMTL methods MaT-FL~\cite{MaT-FL} and FedMTL~\cite{FedMTL}.  

It is noteworthy that PCAGrad~\cite{PCGrad} and CAGrad~\cite{CAGrad} are adapted for multi-task optimization using accumulated gradients transmitted by clients during FL communication rounds. Besides, FedRep~\cite{fedrep} and MaT-FL~\cite{MaT-FL} employ a parameter decoupling strategy as~\cref{eq:prameter decouple} for optimization. By default, the model’s encoder serves as a feature extractor, and only this portion of the parameters is transmitted during the FL process. Furthermore, in the NIID-6 UBCDMT scenario from~\cref{tab: Comp Exp NIID-6 UBCDMT}, due to the diversity in the number and types of tasks among different clients, the model based on the MD architecture exhibits heterogeneity. Consequently, for this scenario, FedProx, FedAMP, PCGrad, CAGrad, and the FedMTL algorithms are modified using parameter decoupling strategy. This implies that algorithms with the ``-E'' flag only transmit and utilize the parameters or accumulated gradient of model encoder for FL optimization.

\subsubsection{Evaluation Criteria}\label{sec:comp. exp Eval}
In our comparative studies, we devise evaluative indicators that comprehensively reflect the unique attributes of individual tasks, the holistic performance of multi-task learning, and the diversity of test set origins.   

\noindent \textbf{Task-Specific Metrics}: Each task type is evaluated using its appropriate metric. For example, depth estimation is assessed using the Root Mean Square Error (RMSE), while the test Loss is used for edge detection. The mean error (mErr) is utilized for surface normal estimation, and the mean Intersection over Union (mIoU) is applied for both semantic segmentation and human parts segmentation.

\noindent\textbf{Comprehensive Multi-Task Performance}: To provide a holistic evaluation of various algorithms, we compute weighted average per-task performance improvement~\cite{astmt}, relative to the target local training baseline without any aggregation. The calculation is now adjusted to incorporate individual task weights:  
\begin{equation}
\Delta\% = \frac{1}{\sum_{i=1}^{N} w_{i}}\sum_{i=1}^{N} (-1)^{l_{i}} w_{i} \frac{M_{\texttt{Fed},i}-M_{\texttt{Target},i}}{M_{\texttt{Target},i}} \times 100\%.
\label{eq:avg perform improvement}
\end{equation}  
In this formula, $N$ signifies the task count, and $M_{\texttt{Fed},i}$ and $M_{\texttt{Target},i}$ denote the performance of task $i$ under federated learning techniques and the target local baseline, respectively. The variable $l_i$ equals 1 when a lower metric value is desirable for task $i$, and 0 otherwise. The weight assigned to task $i$, represented by $w_i$, denotes the importance or priority of each task within the FMTL framework. By default, all tasks are assigned equal weight, i.e., $w_i = 1/N, \forall i$.

\begin{table*}[htbp] 
\caption{Comparative Experiments: Single-Domain Multi-Task (SDMT) and Single-Domain Hybrid-Task (SDHT) Scenarios.\quad \quad}
\vspace{-10pt}
\begin{flushleft}
\textbf{Above}: SDMT (IID-1) with four Multi-Task (MT) clients using NYUD-v2. \textbf{Below}: SDHT (NIID-3) with four MT and four single-task clients using NYUD-v2.   
\textbf{Notations}: `BN' denotes Backbone Network, `Algo' represents Optimization Algorithm, `Ar' signifies Architecture, and `OOM' is Out of Memory. G-FL and P-FL refer to evaluations using global and local test sets, respectively. `$\uparrow$' indicates higher is better, `$\downarrow$' implies lower is better. An asterisk (*) means actual results to be multiplied by $1E-2$. Values before and after `±' represent the average and standard deviation of performance indicators from multiple clients. Light blue shading indicates a baseline for comparison. `$\Delta\%$' denotes average per-task performance improvement relative to the target baseline.  
\textbf{Unless otherwise noted, subsequent tables share these notations.} 
Please refer to \cref{fig: comp. exp overview} for relationship diagrams and training data distribution visualizations across comparative experiments. 
\end{flushleft}
\small
\centering
\setlength{\tabcolsep}{3.5pt} 
\resizebox{\linewidth}{!}{
\begin{tabular}{@{}c|c|l|c|ccccc|ccccc@{}}
\bottomrule
                           &                           & \multicolumn{1}{c|}{}                       &                                              & \multicolumn{5}{c|}{G-FL}                                                                                                                                                                               & \multicolumn{5}{c}{P-FL}                                                                                                                                                            \\ \cline{5-14} 
                           &                           & \multicolumn{1}{c|}{}                       &                                              & Depth                               & Edge                              & Normals                            & \multicolumn{1}{c|}{Semseg}                             &                                & Depth                                & Edge                              & Normals                            & Semseg                             &                                \\
\multirow{-3}{*}{Scenario} & \multirow{-3}{*}{BN}      & \multicolumn{1}{c|}{\multirow{-3}{*}{Algo}} & \multirow{-3}{*}{Ar}                         & RSME*↓                              & Loss*↓                            & mErr↓                              & \multicolumn{1}{c|}{mIoU↑}                              & \multirow{-2}{*}{${\Delta_\textsc{G}\%\uparrow}$}         & RSME*↓                               & Loss*↓                            & mErr↓                              & mIoU↑                              & \multirow{-2}{*}{${\Delta_\textsc{P}\%\uparrow}$}         \\
\end{tabular}
}
\label{tab: Comp Exp IID-1 and NIID-3}
\end{table*}

\begin{table}[tbp]
\setlength{\tabcolsep}{1pt} 
\centering
\caption{Comparative Experiment: NIID-2 Single-Domain Single-Task (SDST) scenario. In this scenario, each client holds data pertaining to distinct task types. Refer to Table \ref{tab: Comp Exp IID-1 and NIID-3} for notation definitions and comparison baseline.} 
\vspace{-10pt}
\resizebox{\linewidth}{!}{
\begin{tabular}{@{}c|l|c|ccccc|ccccc@{}}
\bottomrule
                          & \multicolumn{1}{c|}{}                       &                            & \multicolumn{5}{c|}{G-FL}                                                                                                                                                           & \multicolumn{5}{c}{P-FL}                                                                                                                                                            \\ \cline{4-13} 
                          & \multicolumn{1}{c|}{}                       &                            & Depth                          & Edge                         & Normals                       & \multicolumn{1}{c|}{Semseg}                        &                                & Depth                          & Edge                         & Normals                       & \multicolumn{1}{c|}{Semseg}                        &                                \\
\multirow{-3}{*}{BN}      & \multicolumn{1}{c|}{\multirow{-3}{*}{Algo}} & \multirow{-3}{*}{Ar}       & RSME*↓                         & Loss*↓                       & mErr↓                         & \multicolumn{1}{c|}{mIoU↑}                         & \multirow{-2}{*}{${\Delta_\textsc{G}\%\uparrow}$}         & RSME*↓                         & Loss*↓                       & mErr↓                         & \multicolumn{1}{c|}{mIoU↑}                         & \multirow{-2}{*}{${\Delta_\textsc{P}\%\uparrow}$}         \\ \hline
                          &                                             & SD                         & 81.13                          & 4.75                         & 26.61                         & \multicolumn{1}{c|}{24.68}                         & 1.39                           & 68.93                          & 4.71                         & 27.64                         & \multicolumn{1}{c|}{23.85}                         & 4.42                           \\
\end{tabular}
}
\vspace{-10pt}
\label{tab: Comp Exp NIID-2 SDST}
\end{table}

\begin{table}[tbp]
\setlength{\tabcolsep}{1pt} 
\centering
\caption{Comparative Experiment: NIID-5 Unbalanced Single-Domain Single-Task (UBSDST) with ResNet backbone. Each client holds distinct task types and variable sample numbers. Notations and the target baseline are as defined in \cref{tab: Comp Exp IID-1 and NIID-3}.}
\vspace{-10pt}
\resizebox{\linewidth}{!}{
\begin{tabular}{@{}l|c|ccccc|ccccc@{}}
\bottomrule
\multicolumn{1}{c|}{}                       &                            & \multicolumn{5}{c|}{G-FL}                                                                                                                                                           & \multicolumn{5}{c}{P-FL}                                                                                                                                                            \\ \cline{3-12} 
\multicolumn{1}{c|}{}                       &                            & Depth                          & Edge                         & Normals                       & \multicolumn{1}{c|}{Semseg}                        &                                & Depth                          & Edge                         & Normals                       & \multicolumn{1}{c|}{Semseg}                        &                                \\
\multicolumn{1}{c|}{\multirow{-3}{*}{Algo}} & \multirow{-3}{*}{Ar}       & RSME*↓                         & Loss*↓                       & mErr↓                         & \multicolumn{1}{c|}{mIoU↑}                         & \multirow{-2}{*}{${\Delta_\textsc{G}\%\uparrow}$}         & RSME*↓                         & Loss*↓                       & mErr↓                         & \multicolumn{1}{c|}{mIoU↑}                         & \multirow{-2}{*}{${\Delta_\textsc{P}\%\uparrow}$}         \\ \hline
                                            & SD                         & 75.48                          & 4.75                         & 27.06                         & \multicolumn{1}{c|}{16.89}                         & -5.64                          & 66.87                          & 4.68                         & 27.93                         & \multicolumn{1}{c|}{13.14}                         & -5.96                          \\
\end{tabular}
}
\vspace{-10pt}
\label{tab: Comp Exp NIID-5 UBSDST}
\end{table}

\begin{table}[tbp]
\setlength{\tabcolsep}{3pt} 
\centering
\caption{Comparative Experiment: NIID-7 Unbalanced Cross-Domain Single-Task (UBCDST) with ResNet backbone.}
\vspace{-10pt}
\begin{flushleft}
    Symbols (P) represent PASCAL-Context dataset. The configuration for first four clients mirrors that of NIID-2 scenario detailed in \cref{tab: Comp Exp NIID-2 SDST}. In addition, Clients 4 and 5 incorporate larger datasets from a different domain. Client 4 introduces a new task, human parts segmentation, while Client 5 undertakes surface normal estimation as Client 1. Light blue shading indicates baseline for comparison.  Refer to Tab. \ref{tab: Comp Exp IID-1 and NIID-3} for other notations.
\end{flushleft}
\footnotesize
\resizebox{\linewidth}{!}{
\begin{tabular}{@{}c|l|c|cccccc|c@{}}
\bottomrule
                        & \multicolumn{1}{c|}{}                       &                            & Depth                       & Edge                      & Normals                    & Semseg                     & Parts(P)                      & Normals(P)                    &                                \\
\multirow{-2}{*}{Eval}  & \multicolumn{1}{c|}{\multirow{-2}{*}{Algo}} & \multirow{-2}{*}{Ar}       & RSME*↓                         & Loss*↓                       & mErr↓                         & mIoU↑                         & mIoU↑                         & mErr↓                         & \multirow{-2}{*}{${\Delta\%\uparrow}$}          \\ \hline
                        &                                             & \cellcolor[HTML]{BDD7EE}SD & \cellcolor[HTML]{BDD7EE}80.38  & \cellcolor[HTML]{BDD7EE}4.75 & \cellcolor[HTML]{BDD7EE}26.55 & \cellcolor[HTML]{BDD7EE}24.64 & \cellcolor[HTML]{BDD7EE}55.01 & \cellcolor[HTML]{BDD7EE}14.57 & \cellcolor[HTML]{BDD7EE}0.00   \\
                        & \multirow{-2}{*}{Local}                     & \cellcolor[HTML]{E7E6E6}TC & \cellcolor[HTML]{E7E6E6}79.27  & \cellcolor[HTML]{E7E6E6}4.75 & \cellcolor[HTML]{E7E6E6}26.62 & \cellcolor[HTML]{E7E6E6}24.62 & \cellcolor[HTML]{E7E6E6}54.80 & \cellcolor[HTML]{E7E6E6}14.59 & \cellcolor[HTML]{E7E6E6}0.09   \\ \cline{2-10} 
                        &                                             & SD                         & 111.71                         & 4.85                         & 30.18                         & 10.63                         & 39.25                         & 17.92                         & -27.21                         \\
                        & \multirow{-2}{*}{FedAvg}                    & \cellcolor[HTML]{E7E6E6}TC & \cellcolor[HTML]{E7E6E6}81.30  & \cellcolor[HTML]{E7E6E6}4.81 & \cellcolor[HTML]{E7E6E6}25.96 & \cellcolor[HTML]{E7E6E6}22.57 & \cellcolor[HTML]{E7E6E6}50.38 & \cellcolor[HTML]{E7E6E6}15.65 & \cellcolor[HTML]{E7E6E6}-4.07  \\ \cline{2-10} 
                        &                                             & SD                         & 111.66                         & 4.85                         & 29.98                         & 10.99                         & 40.03                         & 17.69                         & -26.33                         \\
                        & \multirow{-2}{*}{FedProx}                   & \cellcolor[HTML]{E7E6E6}TC & \cellcolor[HTML]{E7E6E6}80.04  & \cellcolor[HTML]{E7E6E6}4.81 & \cellcolor[HTML]{E7E6E6}25.95 & \cellcolor[HTML]{E7E6E6}22.55 & \cellcolor[HTML]{E7E6E6}50.45 & \cellcolor[HTML]{E7E6E6}15.63 & \cellcolor[HTML]{E7E6E6}-3.77  \\ \cline{2-10} 
                        &                                             & SD                         & 80.42                          & 4.75                         & 26.55                         & 24.77                         & 55.07                         & 14.58                         & 0.09                           \\
                        & \multirow{-2}{*}{FedAMP}                    & \cellcolor[HTML]{E7E6E6}TC & \cellcolor[HTML]{E7E6E6}79.60  & \cellcolor[HTML]{E7E6E6}4.75 & \cellcolor[HTML]{E7E6E6}26.61 & \cellcolor[HTML]{E7E6E6}24.19 & \cellcolor[HTML]{E7E6E6}54.78 & \cellcolor[HTML]{E7E6E6}14.60 & \cellcolor[HTML]{E7E6E6}-0.28  \\ \cline{2-10} 
                        &                                             & SD                         & 80.79                          & 4.75                         & 26.59                         & 23.49                         & 52.47                         & 15.00                         & -2.15                          \\
                        & \multirow{-2}{*}{MaT-FL}                    & \cellcolor[HTML]{E7E6E6}TC & \cellcolor[HTML]{E7E6E6}80.97  & \cellcolor[HTML]{E7E6E6}4.75 & \cellcolor[HTML]{E7E6E6}26.56 & \cellcolor[HTML]{E7E6E6}23.27 & \cellcolor[HTML]{E7E6E6}52.94 & \cellcolor[HTML]{E7E6E6}14.91 & \cellcolor[HTML]{E7E6E6}-2.07  \\ \cline{2-10} 
                        &                                             & SD                         & 113.39                         & 4.86                         & 31.13                         & 8.49                          & 34.67                         & 19.07                         & -32.34                         \\
                        & \multirow{-2}{*}{PCGrad}                    & \cellcolor[HTML]{E7E6E6}TC & \cellcolor[HTML]{E7E6E6}117.70 & \cellcolor[HTML]{E7E6E6}4.85 & \cellcolor[HTML]{E7E6E6}30.69 & \cellcolor[HTML]{E7E6E6}15.57 & \cellcolor[HTML]{E7E6E6}39.24 & \cellcolor[HTML]{E7E6E6}18.34 & \cellcolor[HTML]{E7E6E6}-25.91 \\ \cline{2-10} 
                        &                                             & SD                         & 111.52                         & 4.84                         & 29.85                         & 10.82                         & 39.71                         & 17.99                         & -26.74                         \\
                        & \multirow{-2}{*}{CAGrad}                    & \cellcolor[HTML]{E7E6E6}TC & \cellcolor[HTML]{E7E6E6}80.72  & \cellcolor[HTML]{E7E6E6}4.82 & \cellcolor[HTML]{E7E6E6}26.63 & \cellcolor[HTML]{E7E6E6}22.16 & \cellcolor[HTML]{E7E6E6}49.37 & \cellcolor[HTML]{E7E6E6}16.03 & \cellcolor[HTML]{E7E6E6}-5.42  \\ \cline{2-10} 
                        &                                             & SD                         & 80.44                          & 4.75                         & 26.62                         & 23.60                         & 52.36                         & 15.02                         & -2.08                          \\
                        & \multirow{-2}{*}{FedRep}                    & \cellcolor[HTML]{E7E6E6}TC & \cellcolor[HTML]{E7E6E6}78.17  & \cellcolor[HTML]{E7E6E6}4.75 & \cellcolor[HTML]{E7E6E6}26.77 & \cellcolor[HTML]{E7E6E6}23.54 & \cellcolor[HTML]{E7E6E6}52.52 & \cellcolor[HTML]{E7E6E6}14.99 & \cellcolor[HTML]{E7E6E6}-1.66  \\ \cline{2-10} 
                        &                                             & SD                         & 80.54                          & 4.74                         & 26.56                         & 24.74                         & 54.97                         & 14.59                         & 0.03                           \\
\multirow{-18}{*}{G-FL} & \multirow{-2}{*}{FedMTL}                    & \cellcolor[HTML]{E7E6E6}TC & \cellcolor[HTML]{E7E6E6}80.03  & \cellcolor[HTML]{E7E6E6}4.75 & \cellcolor[HTML]{E7E6E6}26.61 & \cellcolor[HTML]{E7E6E6}23.89 & \cellcolor[HTML]{E7E6E6}54.68 & \cellcolor[HTML]{E7E6E6}14.60 & \cellcolor[HTML]{E7E6E6}-0.61  \\ \hline
                        &                                             & \cellcolor[HTML]{BDD7EE}SD & \cellcolor[HTML]{BDD7EE}69.08  & \cellcolor[HTML]{BDD7EE}4.71 & \cellcolor[HTML]{BDD7EE}27.14 & \cellcolor[HTML]{BDD7EE}22.44 & \cellcolor[HTML]{BDD7EE}54.28 & \cellcolor[HTML]{BDD7EE}14.24 & \cellcolor[HTML]{BDD7EE}0.00   \\
                        & \multirow{-2}{*}{Local}                     & \cellcolor[HTML]{E7E6E6}TC & \cellcolor[HTML]{E7E6E6}75.83  & \cellcolor[HTML]{E7E6E6}4.71 & \cellcolor[HTML]{E7E6E6}27.21 & \cellcolor[HTML]{E7E6E6}22.90 & \cellcolor[HTML]{E7E6E6}55.62 & \cellcolor[HTML]{E7E6E6}14.17 & \cellcolor[HTML]{E7E6E6}-0.84  \\ \cline{2-10} 
                        &                                             & SD                         & 103.69                         & 4.82                         & 31.81                         & 9.43                          & 38.11                         & 17.45                         & -29.99                         \\
                        & \multirow{-2}{*}{FedAvg}                    & \cellcolor[HTML]{E7E6E6}TC & \cellcolor[HTML]{E7E6E6}74.99  & \cellcolor[HTML]{E7E6E6}4.77 & \cellcolor[HTML]{E7E6E6}27.12 & \cellcolor[HTML]{E7E6E6}20.67 & \cellcolor[HTML]{E7E6E6}51.12 & \cellcolor[HTML]{E7E6E6}15.33 & \cellcolor[HTML]{E7E6E6}-5.19  \\ \cline{2-10} 
                        &                                             & SD                         & 102.58                         & 4.83                         & 31.71                         & 10.07                         & 38.50                         & 17.30                         & -28.93                         \\
                        & \multirow{-2}{*}{FedProx}                   & \cellcolor[HTML]{E7E6E6}TC & \cellcolor[HTML]{E7E6E6}73.09  & \cellcolor[HTML]{E7E6E6}4.77 & \cellcolor[HTML]{E7E6E6}27.17 & \cellcolor[HTML]{E7E6E6}19.61 & \cellcolor[HTML]{E7E6E6}49.88 & \cellcolor[HTML]{E7E6E6}15.33 & \cellcolor[HTML]{E7E6E6}-5.93  \\ \cline{2-10} 
                        &                                             & SD                         & 71.89                          & 4.71                         & 27.82                         & 24.44                         & 55.65                         & 14.14                         & 0.93                           \\
                        & \multirow{-2}{*}{FedAMP}                    & \cellcolor[HTML]{E7E6E6}TC & \cellcolor[HTML]{E7E6E6}69.65  & \cellcolor[HTML]{E7E6E6}4.70 & \cellcolor[HTML]{E7E6E6}27.28 & \cellcolor[HTML]{E7E6E6}22.40 & \cellcolor[HTML]{E7E6E6}55.16 & \cellcolor[HTML]{E7E6E6}14.15 & \cellcolor[HTML]{E7E6E6}0.16   \\ \cline{2-10} 
                        &                                             & SD                         & 71.26                          & 4.71                         & 27.79                         & 20.16                         & 53.21                         & 14.62                         & -3.39                          \\
                        & \multirow{-2}{*}{MaT-FL}                    & \cellcolor[HTML]{E7E6E6}TC & \cellcolor[HTML]{E7E6E6}77.87  & \cellcolor[HTML]{E7E6E6}4.70 & \cellcolor[HTML]{E7E6E6}28.17 & \cellcolor[HTML]{E7E6E6}20.75 & \cellcolor[HTML]{E7E6E6}52.95 & \cellcolor[HTML]{E7E6E6}14.50 & \cellcolor[HTML]{E7E6E6}-4.69  \\ \cline{2-10} 
                        &                                             & SD                         & 107.86                         & 4.83                         & 32.36                         & 8.84                          & 34.36                         & 18.91                         & -34.67                         \\
                        & \multirow{-2}{*}{PCGrad}                    & \cellcolor[HTML]{E7E6E6}TC & \cellcolor[HTML]{E7E6E6}105.01 & \cellcolor[HTML]{E7E6E6}4.80 & \cellcolor[HTML]{E7E6E6}32.40 & \cellcolor[HTML]{E7E6E6}14.79 & \cellcolor[HTML]{E7E6E6}37.20 & \cellcolor[HTML]{E7E6E6}18.03 & \cellcolor[HTML]{E7E6E6}-27.58 \\ \cline{2-10} 
                        &                                             & SD                         & 102.67                         & 4.81                         & 32.04                         & 10.01                         & 37.76                         & 17.51                         & -29.60                         \\
                        & \multirow{-2}{*}{CAGrad}                    & \cellcolor[HTML]{E7E6E6}TC & \cellcolor[HTML]{E7E6E6}69.89  & \cellcolor[HTML]{E7E6E6}4.78 & \cellcolor[HTML]{E7E6E6}28.03 & \cellcolor[HTML]{E7E6E6}19.23 & \cellcolor[HTML]{E7E6E6}48.88 & \cellcolor[HTML]{E7E6E6}15.67 & \cellcolor[HTML]{E7E6E6}-6.71  \\ \cline{2-10} 
                        &                                             & SD                         & 71.55                          & 4.71                         & 28.01                         & 21.04                         & 52.51                         & 14.65                         & -3.19                          \\
                        & \multirow{-2}{*}{FedRep}                    & \cellcolor[HTML]{E7E6E6}TC & \cellcolor[HTML]{E7E6E6}74.10  & \cellcolor[HTML]{E7E6E6}4.71 & \cellcolor[HTML]{E7E6E6}28.34 & \cellcolor[HTML]{E7E6E6}22.58 & \cellcolor[HTML]{E7E6E6}53.08 & \cellcolor[HTML]{E7E6E6}14.54 & \cellcolor[HTML]{E7E6E6}-2.56  \\ \cline{2-10} 
                        &                                             & SD                         & 67.38                          & 4.71                         & 27.29                         & 22.35                         & 54.50                         & 14.18                         & 0.39                           \\
\multirow{-18}{*}{P-FL} & \multirow{-2}{*}{FedMTL}                    & \cellcolor[HTML]{E7E6E6}TC & \cellcolor[HTML]{E7E6E6}69.90  & \cellcolor[HTML]{E7E6E6}4.72 & \cellcolor[HTML]{E7E6E6}27.54 & \cellcolor[HTML]{E7E6E6}22.71 & \cellcolor[HTML]{E7E6E6}54.55 & \cellcolor[HTML]{E7E6E6}14.12 & \cellcolor[HTML]{E7E6E6}-0.05  \\ \toprule
\end{tabular}
}
\label{tab: Comp Exp NIID-7 UBCDST}
\end{table}

\begin{table*}[tbp]
\setlength{\tabcolsep}{8pt} 
\centering
\caption{Comparative Experiment: NIID-4 Unbalanced Single-Domain Multi-Task (UBSDMT) scenario. This scenario mirrors the IID-1 as detailed in \cref{tab: Comp Exp IID-1 and NIID-3}, with the sole distinction being an unbalanced distribution of total samples across different clients.}
\vspace{-10pt}
\footnotesize
\resizebox{1\linewidth}{!}{
\begin{tabular}{@{}l|l|ccccc|ccccc@{}}
\bottomrule
\multicolumn{1}{c|}{}                       & \multicolumn{1}{c|}{}                     & \multicolumn{5}{c|}{G-FL}                                                                                                                                                                                & \multicolumn{5}{c}{P-FL}                                                                                                                                                                                 \\ \cline{3-12} 
\multicolumn{1}{c|}{}                       & \multicolumn{1}{c|}{}                     & Depth                                & Edge                              & Normals                            & \multicolumn{1}{c|}{Semseg}                             &                                & Depth                                & Edge                              & Normals                            & \multicolumn{1}{c|}{Semseg}                             &                                \\
\multicolumn{1}{c|}{\multirow{-3}{*}{Algo}} & \multicolumn{1}{c|}{\multirow{-3}{*}{Ar}} & RSME*↓                               & Loss*↓                            & mErr↓                              & \multicolumn{1}{c|}{mIoU↑}                              & \multirow{-2}{*}{${\Delta_\textsc{G}\%\uparrow}$}         & RSME*↓                               & Loss*↓                            & mErr↓                              & \multicolumn{1}{c|}{mIoU↑}                              & \multirow{-2}{*}{${\Delta_\textsc{P}\%\uparrow}$}         \\ \hline
                                            & MD                                        & 84.82±9.99                           & 4.77±0.03                         & 26.82±1.68                         & \multicolumn{1}{c|}{22.16±4.48}                         & -2.74                          & 91.08±20.63                          & 4.81±0.11                         & 26.70±2.52                         & \multicolumn{1}{c|}{22.32±6.16}                         & -2.83                          \\
\multirow{-2}{*}{Local}                     & \cellcolor[HTML]{E7E6E6}TC                & \cellcolor[HTML]{E7E6E6}82.36±7.60   & \cellcolor[HTML]{E7E6E6}4.78±0.00 & \cellcolor[HTML]{E7E6E6}26.72±1.68 & \multicolumn{1}{c|}{\cellcolor[HTML]{E7E6E6}22.04±4.66} & \cellcolor[HTML]{E7E6E6}-2.07  & \cellcolor[HTML]{E7E6E6}87.42±18.66  & \cellcolor[HTML]{E7E6E6}4.82±0.10 & \cellcolor[HTML]{E7E6E6}26.67±2.54 & \multicolumn{1}{c|}{\cellcolor[HTML]{E7E6E6}22.48±6.57} & \cellcolor[HTML]{E7E6E6}-1.69  \\ \hline
                                            & MD                                        & 71.55±0.07                           & 4.77±0.00                         & 22.96±0.01                         & \multicolumn{1}{c|}{29.98±0.10}                         & 13.35                          & 68.15±4.99                           & 4.80±0.10                         & 22.69±0.56                         & \multicolumn{1}{c|}{29.15±4.67}                         & 14.21                          \\
\multirow{-2}{*}{FedAvg}                    & \cellcolor[HTML]{E7E6E6}TC                & \cellcolor[HTML]{E7E6E6}74.57±1.25   & \cellcolor[HTML]{E7E6E6}4.82±0.02 & \cellcolor[HTML]{E7E6E6}22.57±0.02 & \multicolumn{1}{c|}{\cellcolor[HTML]{E7E6E6}30.48±0.45} & \cellcolor[HTML]{E7E6E6}13.06  & \cellcolor[HTML]{E7E6E6}69.68±4.83   & \cellcolor[HTML]{E7E6E6}4.85±0.10 & \cellcolor[HTML]{E7E6E6}22.16±0.47 & \multicolumn{1}{c|}{\cellcolor[HTML]{E7E6E6}26.76±4.00} & \cellcolor[HTML]{E7E6E6}11.62  \\ \hline
                                            & MD                                        & 71.55±0.12                           & 4.77±0.00                         & 22.95±0.02                         & \multicolumn{1}{c|}{29.79±0.09}                         & 13.16                          & 67.39±4.93                           & 4.80±0.10                         & 22.50±0.46                         & \multicolumn{1}{c|}{28.45±3.87}                         & 13.89                          \\
\multirow{-2}{*}{FedProx}                   & \cellcolor[HTML]{E7E6E6}TC                & \cellcolor[HTML]{E7E6E6}72.68±1.18   & \cellcolor[HTML]{E7E6E6}4.82±0.02 & \cellcolor[HTML]{E7E6E6}22.56±0.02 & \multicolumn{1}{c|}{\cellcolor[HTML]{E7E6E6}30.23±0.45} & \cellcolor[HTML]{E7E6E6}13.39  & \cellcolor[HTML]{E7E6E6}70.78±7.83   & \cellcolor[HTML]{E7E6E6}4.85±0.10 & \cellcolor[HTML]{E7E6E6}22.34±0.48 & \multicolumn{1}{c|}{\cellcolor[HTML]{E7E6E6}26.94±3.07} & \cellcolor[HTML]{E7E6E6}11.33  \\ \hline
                                            & MD                                        & 84.86±10.09                          & 4.77±0.03                         & 26.85±1.68                         & \multicolumn{1}{c|}{22.00±4.41}                         & -2.96                          & 91.62±22.57                          & 4.81±0.11                         & 26.38±2.62                         & \multicolumn{1}{c|}{22.62±6.60}                         & -2.37                          \\
\multirow{-2}{*}{FedAMP}                    & \cellcolor[HTML]{E7E6E6}TC                & \cellcolor[HTML]{E7E6E6}83.11±7.45   & \cellcolor[HTML]{E7E6E6}4.78±0.01 & \cellcolor[HTML]{E7E6E6}26.83±1.76 & \multicolumn{1}{c|}{\cellcolor[HTML]{E7E6E6}21.92±4.24} & \cellcolor[HTML]{E7E6E6}-2.54  & \cellcolor[HTML]{E7E6E6}85.32±14.26  & \cellcolor[HTML]{E7E6E6}4.81±0.10 & \cellcolor[HTML]{E7E6E6}26.78±2.70 & \multicolumn{1}{c|}{\cellcolor[HTML]{E7E6E6}21.78±5.42} & \cellcolor[HTML]{E7E6E6}-1.87  \\ \hline
                                            & MD                                        & 82.68±8.21                           & 4.77±0.02                         & 26.02±1.02                         & \multicolumn{1}{c|}{22.61±3.58}                         & -0.85                          & 86.93±18.93                          & 4.80±0.11                         & 25.70±1.66                         & \multicolumn{1}{c|}{23.31±5.86}                         & 0.31                           \\
\multirow{-2}{*}{MaT-FL}                    & \cellcolor[HTML]{E7E6E6}TC                & \cellcolor[HTML]{E7E6E6}78.31±4.16   & \cellcolor[HTML]{E7E6E6}4.78±0.01 & \cellcolor[HTML]{E7E6E6}25.62±0.89 & \multicolumn{1}{c|}{\cellcolor[HTML]{E7E6E6}23.94±3.62} & \cellcolor[HTML]{E7E6E6}2.24   & \cellcolor[HTML]{E7E6E6}82.91±13.85  & \cellcolor[HTML]{E7E6E6}4.82±0.11 & \cellcolor[HTML]{E7E6E6}25.28±1.42 & \multicolumn{1}{c|}{\cellcolor[HTML]{E7E6E6}23.25±5.28} & \cellcolor[HTML]{E7E6E6}1.65   \\ \hline
                                            & MD                                        & 141.20±11.51                         & 4.86±0.01                         & 29.04±0.88                         & \multicolumn{1}{c|}{14.55±1.27}                         & -30.77                         & 146.40±18.90                         & 4.89±0.10                         & 28.25±1.09                         & \multicolumn{1}{c|}{14.81±2.01}                         & -27.46                         \\
\multirow{-2}{*}{PCGrad}                    & \cellcolor[HTML]{E7E6E6}TC                & \cellcolor[HTML]{E7E6E6}105.72±15.51 & \cellcolor[HTML]{E7E6E6}4.95±0.04 & \cellcolor[HTML]{E7E6E6}32.33±0.59 & \multicolumn{1}{c|}{\cellcolor[HTML]{E7E6E6}17.24±2.54} & \cellcolor[HTML]{E7E6E6}-20.58 & \cellcolor[HTML]{E7E6E6}107.12±17.75 & \cellcolor[HTML]{E7E6E6}4.99±0.09 & \cellcolor[HTML]{E7E6E6}32.43±1.37 & \multicolumn{1}{c|}{\cellcolor[HTML]{E7E6E6}16.53±3.74} & \cellcolor[HTML]{E7E6E6}-19.45 \\ \hline
                                            & MD                                        & 84.53±6.03                           & 4.81±0.01                         & 25.33±0.87                         & \multicolumn{1}{c|}{19.66±2.36}                         & -4.13                          & 86.12±15.21                          & 4.85±0.10                         & 25.14±1.24                         & \multicolumn{1}{c|}{20.51±3.80}                         & -2.03                          \\
\multirow{-2}{*}{CAGrad}                    & \cellcolor[HTML]{E7E6E6}TC                & \cellcolor[HTML]{E7E6E6}79.71±3.73   & \cellcolor[HTML]{E7E6E6}4.86±0.01 & \cellcolor[HTML]{E7E6E6}24.48±0.82 & \multicolumn{1}{c|}{\cellcolor[HTML]{E7E6E6}23.59±3.29} & \cellcolor[HTML]{E7E6E6}2.10   & \cellcolor[HTML]{E7E6E6}78.93±6.44   & \cellcolor[HTML]{E7E6E6}4.89±0.10 & \cellcolor[HTML]{E7E6E6}24.20±1.20 & \multicolumn{1}{c|}{\cellcolor[HTML]{E7E6E6}23.18±5.55} & \cellcolor[HTML]{E7E6E6}3.32   \\ \hline
                                            & MD                                        & 81.55±8.21                           & 4.77±0.02                         & 25.62±1.08                         & \multicolumn{1}{c|}{23.03±3.63}                         & 0.33                           & 84.27±17.17                          & 4.80±0.11                         & 25.50±1.81                         & \multicolumn{1}{c|}{23.11±5.93}                         & 1.03                           \\
\multirow{-2}{*}{FedRep}                    & \cellcolor[HTML]{E7E6E6}TC                & \cellcolor[HTML]{E7E6E6}77.41±5.02   & \cellcolor[HTML]{E7E6E6}4.78±0.01 & \cellcolor[HTML]{E7E6E6}24.93±0.95 & \multicolumn{1}{c|}{\cellcolor[HTML]{E7E6E6}24.67±4.05} & \cellcolor[HTML]{E7E6E6}3.95   & \cellcolor[HTML]{E7E6E6}75.64±10.29  & \cellcolor[HTML]{E7E6E6}4.82±0.10 & \cellcolor[HTML]{E7E6E6}24.54±1.63 & \multicolumn{1}{c|}{\cellcolor[HTML]{E7E6E6}24.00±5.49} & \cellcolor[HTML]{E7E6E6}5.10   \\ \hline
                                            & MD                                        & 85.10±9.98                           & 4.77±0.02                         & 26.85±1.66                         & \multicolumn{1}{c|}{21.92±4.32}                         & -3.11                          & 90.05±19.50                          & 4.80±0.11                         & 26.46±2.49                         & \multicolumn{1}{c|}{23.29±5.93}                         & -1.28                          \\
\multirow{-2}{*}{FedMTL}                    & \cellcolor[HTML]{E7E6E6}TC                & \cellcolor[HTML]{E7E6E6}83.21±7.25   & \cellcolor[HTML]{E7E6E6}4.78±0.01 & \cellcolor[HTML]{E7E6E6}26.76±1.82 & \multicolumn{1}{c|}{\cellcolor[HTML]{E7E6E6}22.22±4.40} & \cellcolor[HTML]{E7E6E6}-2.18  & \cellcolor[HTML]{E7E6E6}86.52±16.15  & \cellcolor[HTML]{E7E6E6}4.82±0.10 & \cellcolor[HTML]{E7E6E6}26.22±2.22 & \multicolumn{1}{c|}{\cellcolor[HTML]{E7E6E6}21.77±6.11} & \cellcolor[HTML]{E7E6E6}-1.73  \\ \toprule
\end{tabular}
\label{tab: Comp Exp NIID-4 UBSDMT}
}
\end{table*}

\begin{table}[tbp]
\setlength{\tabcolsep}{2pt} 
\centering
\caption{Comparative Experiment: NIID-6 Unbalanced Cross-Domain Multi-Task (UBCDMT) with global evaluation G-FL.}
\vspace{-10pt}
\begin{flushleft}
    The setup for the first four clients mirrors that of the IID-1 scenario as detailed in Table \ref{tab: Comp Exp IID-1 and NIID-3}. Client 4 incorporates a larger dataset from a different domain, introducing a new task of human parts segmentation while also performing the same surface normal estimation task as Client 1. `NULL' indicates the absence of such a baseline. The `-E' flag is used when only the parameters of the model encoder or the accumulated gradient are transmitted and utilized for federated learning optimization. For additional notation definitions, please refer to Table \ref{tab: Comp Exp IID-1 and NIID-3} and Table \ref{tab: Comp Exp NIID-7 UBCDST}.
\end{flushleft}
\resizebox{\linewidth}{!}{
\begin{tabular}{@{}c|l|c|cccccc|c@{}}
\bottomrule
                          & \multicolumn{1}{c|}{}                       &                            & Depth                            & Edge                           & Normals                         & Semseg                          & Parts(P)                      & Normals(P)                    &                                \\
\multirow{-2}{*}{BN}      & \multicolumn{1}{c|}{\multirow{-2}{*}{Algo}} & \multirow{-2}{*}{Ar}       & RSME*↓                              & Loss*↓                            & mErr↓                              & mIoU↑                              & mIoU↑                         & mErr↓                         & \multirow{-2}{*}{${\Delta_\textsc{G}\%\uparrow}$}         \\ \hline
                          & Local                                       & \cellcolor[HTML]{BDD7EE}MD & \cellcolor[HTML]{BDD7EE}81.82±2.09  & \cellcolor[HTML]{BDD7EE}4.76±0.00 & \cellcolor[HTML]{BDD7EE}26.49±0.13 & \cellcolor[HTML]{BDD7EE}23.37±0.64 & \cellcolor[HTML]{BDD7EE}54.12 & \cellcolor[HTML]{BDD7EE}13.86 & \cellcolor[HTML]{BDD7EE}0.00   \\
                          & Local                                       & \cellcolor[HTML]{E7E6E6}TC & \cellcolor[HTML]{E7E6E6}80.44±1.86  & \cellcolor[HTML]{E7E6E6}4.78±0.01 & \cellcolor[HTML]{E7E6E6}26.43±0.17 & \cellcolor[HTML]{E7E6E6}22.72±0.40 & \cellcolor[HTML]{E7E6E6}52.54 & \cellcolor[HTML]{E7E6E6}13.78 & \cellcolor[HTML]{E7E6E6}-0.61  \\ \cline{2-10} 
                          & FedAvg                                      & MD                         & NULL                                & NULL                              & NULL                               & NULL                               & NULL                          & NULL                          & NULL                           \\
                          & FedAvg                                      & \cellcolor[HTML]{E7E6E6}TC & \cellcolor[HTML]{E7E6E6}75.27±0.20  & \cellcolor[HTML]{E7E6E6}4.82±0.00 & \cellcolor[HTML]{E7E6E6}22.15±0.04 & \cellcolor[HTML]{E7E6E6}30.50±0.19 & \cellcolor[HTML]{E7E6E6}50.59 & \cellcolor[HTML]{E7E6E6}14.89 & \cellcolor[HTML]{E7E6E6}6.61   \\ \cline{2-10} 
                          & FedProx-E                                   & MD                         & 78.55±1.17                          & 4.76±0.00                         & 25.14±0.20                         & 23.61±0.55                         & 54.08                         & 14.11                         & 1.37                           \\
                          & FedProx                                     & \cellcolor[HTML]{E7E6E6}TC & \cellcolor[HTML]{E7E6E6}73.76±0.20  & \cellcolor[HTML]{E7E6E6}4.82±0.00 & \cellcolor[HTML]{E7E6E6}22.23±0.04 & \cellcolor[HTML]{E7E6E6}30.06±0.19 & \cellcolor[HTML]{E7E6E6}50.22 & \cellcolor[HTML]{E7E6E6}14.88 & \cellcolor[HTML]{E7E6E6}6.46   \\ \cline{2-10} 
                          & FedAMP-E                                    & MD                         & 81.33±1.74                          & 4.76±0.00                         & 26.50±0.13                         & 23.16±0.37                         & 54.2                          & 13.87                         & -0.04                          \\
                          & FedAMP                                      & \cellcolor[HTML]{E7E6E6}TC & \cellcolor[HTML]{E7E6E6}80.49±1.58  & \cellcolor[HTML]{E7E6E6}4.78±0.01 & \cellcolor[HTML]{E7E6E6}26.47±0.16 & \cellcolor[HTML]{E7E6E6}22.48±0.54 & \cellcolor[HTML]{E7E6E6}53.05 & \cellcolor[HTML]{E7E6E6}13.84 & \cellcolor[HTML]{E7E6E6}-0.73  \\ \cline{2-10} 
                          & MaT-FL                                      & MD                         & 79.33±0.92                          & 4.76±0.00                         & 25.48±0.48                         & 23.24±0.51                         & 54.21                         & 14.06                         & 0.84                           \\
                          & MaT-FL                                      & \cellcolor[HTML]{E7E6E6}TC & \cellcolor[HTML]{E7E6E6}78.44±3.28  & \cellcolor[HTML]{E7E6E6}4.79±0.01 & \cellcolor[HTML]{E7E6E6}25.24±0.77 & \cellcolor[HTML]{E7E6E6}23.78±0.72 & \cellcolor[HTML]{E7E6E6}53.19 & \cellcolor[HTML]{E7E6E6}14.01 & \cellcolor[HTML]{E7E6E6}1.20   \\ \cline{2-10} 
                          & PCGrad-E                                    & MD                         & 88.96±1.27                          & 4.79±0.00                         & 30.43±0.28                         & 16.32±0.40                         & 46.73                         & 18.67                         & -17.13                         \\
                          & PCGrad                                      & \cellcolor[HTML]{E7E6E6}TC & \cellcolor[HTML]{E7E6E6}96.21±1.95  & \cellcolor[HTML]{E7E6E6}4.94±0.02 & \cellcolor[HTML]{E7E6E6}32.90±0.43 & \cellcolor[HTML]{E7E6E6}16.86±0.47 & \cellcolor[HTML]{E7E6E6}43.75 & \cellcolor[HTML]{E7E6E6}24.84 & \cellcolor[HTML]{E7E6E6}-28.63 \\ \cline{2-10} 
                          & CAGrad-E                                    & MD                         & 78.22±1.21                          & 4.75±0.00                         & 25.02±0.16                         & 24.07±0.68                         & 53.78                         & 14.16                         & 1.73                           \\
                          & CAGrad                                      & \cellcolor[HTML]{E7E6E6}TC & \cellcolor[HTML]{E7E6E6}78.71±1.80  & \cellcolor[HTML]{E7E6E6}4.86±0.01 & \cellcolor[HTML]{E7E6E6}23.81±0.04 & \cellcolor[HTML]{E7E6E6}23.95±0.39 & \cellcolor[HTML]{E7E6E6}49.77 & \cellcolor[HTML]{E7E6E6}14.99 & \cellcolor[HTML]{E7E6E6}-0.32  \\ \cline{2-10} 
                          & FedRep                                      & MD                         & 78.86±1.43                          & 4.76±0.00                         & 25.16±0.13                         & 23.66±0.53                         & 54.17                         & 14.08                         & 1.40                           \\
                          & FedRep                                      & \cellcolor[HTML]{E7E6E6}TC & \cellcolor[HTML]{E7E6E6}75.41±1.38  & \cellcolor[HTML]{E7E6E6}4.78±0.01 & \cellcolor[HTML]{E7E6E6}24.52±0.09 & \cellcolor[HTML]{E7E6E6}24.94±0.48 & \cellcolor[HTML]{E7E6E6}52.87 & \cellcolor[HTML]{E7E6E6}14.05 & \cellcolor[HTML]{E7E6E6}2.98   \\ \cline{2-10} 
                          & FedMTL-E                                    & MD                         & 82.02±1.96                          & 4.76±0.00                         & 26.50±0.12                         & 23.14±0.30                         & 53.84                         & 13.86                         & -0.30                          \\
\multirow{-18}{*}{resnet} & FedMTL                                      & \cellcolor[HTML]{E7E6E6}TC & \cellcolor[HTML]{E7E6E6}80.61±1.38  & \cellcolor[HTML]{E7E6E6}4.78±0.01 & \cellcolor[HTML]{E7E6E6}26.43±0.18 & \cellcolor[HTML]{E7E6E6}22.88±0.61 & \cellcolor[HTML]{E7E6E6}52.69 & \cellcolor[HTML]{E7E6E6}13.81 & \cellcolor[HTML]{E7E6E6}-0.52  \\ \hline
                          & Local                                       & MD                         & 72.34±1.14                          & 4.73±0.01                         & 24.4±0.13                          & 33.83±0.38                         & 54.42                         & 13.67                         & 11.13                          \\
                          & Local                                       & \cellcolor[HTML]{E7E6E6}TC & \cellcolor[HTML]{E7E6E6}77.50±2.22  & \cellcolor[HTML]{E7E6E6}4.75±0.00 & \cellcolor[HTML]{E7E6E6}25.19±0.20 & \cellcolor[HTML]{E7E6E6}29.98±0.59 & \cellcolor[HTML]{E7E6E6}52.35 & \cellcolor[HTML]{E7E6E6}13.65 & \cellcolor[HTML]{E7E6E6}6.15   \\ \cline{2-10} 
                          & FedAvg                                      & MD                         & NULL                                & NULL                              & NULL                               & NULL                               & NULL                          & NULL                          & NULL                           \\
                          & FedAvg                                      & \cellcolor[HTML]{E7E6E6}TC & \cellcolor[HTML]{E7E6E6}70.15±0.17  & \cellcolor[HTML]{E7E6E6}4.78±0.00 & \cellcolor[HTML]{E7E6E6}20.96±0.02 & \cellcolor[HTML]{E7E6E6}40.65±0.03 & \cellcolor[HTML]{E7E6E6}52.32 & \cellcolor[HTML]{E7E6E6}14.61 & \cellcolor[HTML]{E7E6E6}16.65  \\ \cline{2-10} 
                          & FedProx-E                                   & MD                         & 70.89±1.19                          & 4.73±0.00                         & 23.92±0.05                         & 33.91±0.37                         & 55.22                         & 14.13                         & 11.48                          \\
                          & FedProx                                     & \cellcolor[HTML]{E7E6E6}TC & \cellcolor[HTML]{E7E6E6}70.12±0.18  & \cellcolor[HTML]{E7E6E6}4.78±0.00 & \cellcolor[HTML]{E7E6E6}20.99±0.02 & \cellcolor[HTML]{E7E6E6}40.51±0.03 & \cellcolor[HTML]{E7E6E6}52.32 & \cellcolor[HTML]{E7E6E6}14.62 & \cellcolor[HTML]{E7E6E6}16.53  \\ \cline{2-10} 
                          & FedAMP-E                                    & MD                         & OOM                                 & OOM                               & OOM                                & OOM                                & OOM                           & OOM                           & OOM                            \\
                          & FedAMP                                      & \cellcolor[HTML]{E7E6E6}TC & \cellcolor[HTML]{E7E6E6}77.77±1.53  & \cellcolor[HTML]{E7E6E6}4.75±0.00 & \cellcolor[HTML]{E7E6E6}25.25±0.15 & \cellcolor[HTML]{E7E6E6}29.95±0.99 & \cellcolor[HTML]{E7E6E6}52.33 & \cellcolor[HTML]{E7E6E6}13.7  & \cellcolor[HTML]{E7E6E6}5.97   \\ \cline{2-10} 
                          & MaT-FL                                      & MD                         & 71.48±0.59                          & 4.73±0.00                         & 23.96±0.19                         & 33.41±0.85                         & 55.42                         & 14.05                         & 11.14                          \\
                          & MaT-FL                                      & \cellcolor[HTML]{E7E6E6}TC & \cellcolor[HTML]{E7E6E6}72.08±1.33  & \cellcolor[HTML]{E7E6E6}4.74±0.00 & \cellcolor[HTML]{E7E6E6}23.80±0.07 & \cellcolor[HTML]{E7E6E6}33.69±0.23 & \cellcolor[HTML]{E7E6E6}52.09 & \cellcolor[HTML]{E7E6E6}13.62 & \cellcolor[HTML]{E7E6E6}10.77  \\ \cline{2-10} 
                          & PCGrad-E                                    & MD                         & 79.38±2.37                          & 4.73±0.00                         & 26.16±0.05                         & 33.25±0.34                         & 52.93                         & 15.38                         & 5.66                           \\
                          & PCGrad                                      & \cellcolor[HTML]{E7E6E6}TC & \cellcolor[HTML]{E7E6E6}85.99±14.50 & \cellcolor[HTML]{E7E6E6}4.90±0.02 & \cellcolor[HTML]{E7E6E6}28.31±0.22 & \cellcolor[HTML]{E7E6E6}34.98±0.49 & \cellcolor[HTML]{E7E6E6}50.84 & \cellcolor[HTML]{E7E6E6}19.68 & \cellcolor[HTML]{E7E6E6}-2.21  \\ \cline{2-10} 
                          & CAGrad-E                                    & MD                         & 69.90±0.82                          & 4.73±0.01                         & 23.67±0.06                         & 34.57±0.33                         & 55                            & 14.21                         & 12.14                          \\
                          & CAGrad                                      & \cellcolor[HTML]{E7E6E6}TC & \cellcolor[HTML]{E7E6E6}75.34±8.49  & \cellcolor[HTML]{E7E6E6}4.81±0.01 & \cellcolor[HTML]{E7E6E6}21.97±0.02 & \cellcolor[HTML]{E7E6E6}34.15±0.86 & \cellcolor[HTML]{E7E6E6}51.1  & \cellcolor[HTML]{E7E6E6}14.69 & \cellcolor[HTML]{E7E6E6}9.75   \\ \cline{2-10} 
                          & FedRep                                      & MD                         & 70.78±1.18                          & 4.73±0.00                         & 23.89±0.04                         & 33.90±0.43                         & 55.32                         & 14.13                         & 11.54                          \\
                          & FedRep                                      & \cellcolor[HTML]{E7E6E6}TC & \cellcolor[HTML]{E7E6E6}72.93±1.22  & \cellcolor[HTML]{E7E6E6}4.75±0.00 & \cellcolor[HTML]{E7E6E6}23.91±0.07 & \cellcolor[HTML]{E7E6E6}31.54±0.58 & \cellcolor[HTML]{E7E6E6}53.01 & \cellcolor[HTML]{E7E6E6}14.2  & \cellcolor[HTML]{E7E6E6}8.55   \\ \cline{2-10} 
                          & FedMTL-E                                    & MD                         & OOM                                 & OOM                               & OOM                                & OOM                                & OOM                           & OOM                           & OOM                            \\
\multirow{-18}{*}{swin-t} & FedMTL                                      & \cellcolor[HTML]{E7E6E6}TC & \cellcolor[HTML]{E7E6E6}77.30±2.03  & \cellcolor[HTML]{E7E6E6}4.75±0.00 & \cellcolor[HTML]{E7E6E6}25.23±0.17 & \cellcolor[HTML]{E7E6E6}29.70±0.54 & \cellcolor[HTML]{E7E6E6}52.32 & \cellcolor[HTML]{E7E6E6}13.64 & \cellcolor[HTML]{E7E6E6}5.97   \\ \toprule
\end{tabular}
}
\label{tab: Comp Exp NIID-6 UBCDMT}
\end{table}

\begin{figure}[tbp]
    \centering
    \includegraphics[width=1\linewidth]{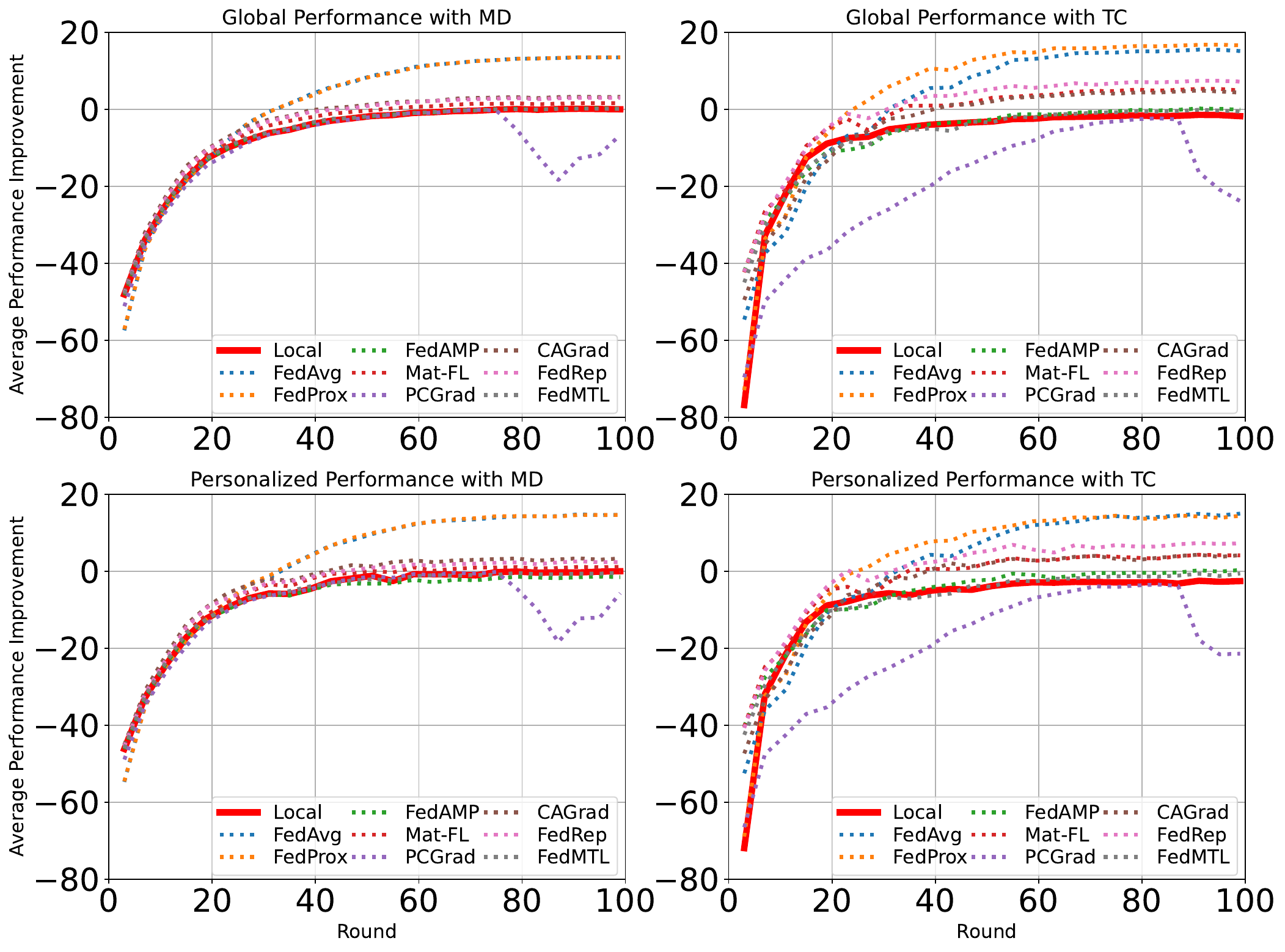}
    \vspace{-10pt}
    \caption{Case Study: Average task performance improvement (${\Delta\%\uparrow}$) versus communication rounds for baselines in IID-1 SDMT scenario from~\cref{tab: Comp Exp IID-1 and NIID-3}. Please zoom in for details.}
    \label{fig:eval IID-1 performance improvement}
    \vspace{-10pt}
\end{figure}

\begin{figure}[tbp]
    \centering
    \includegraphics[width=1\linewidth]{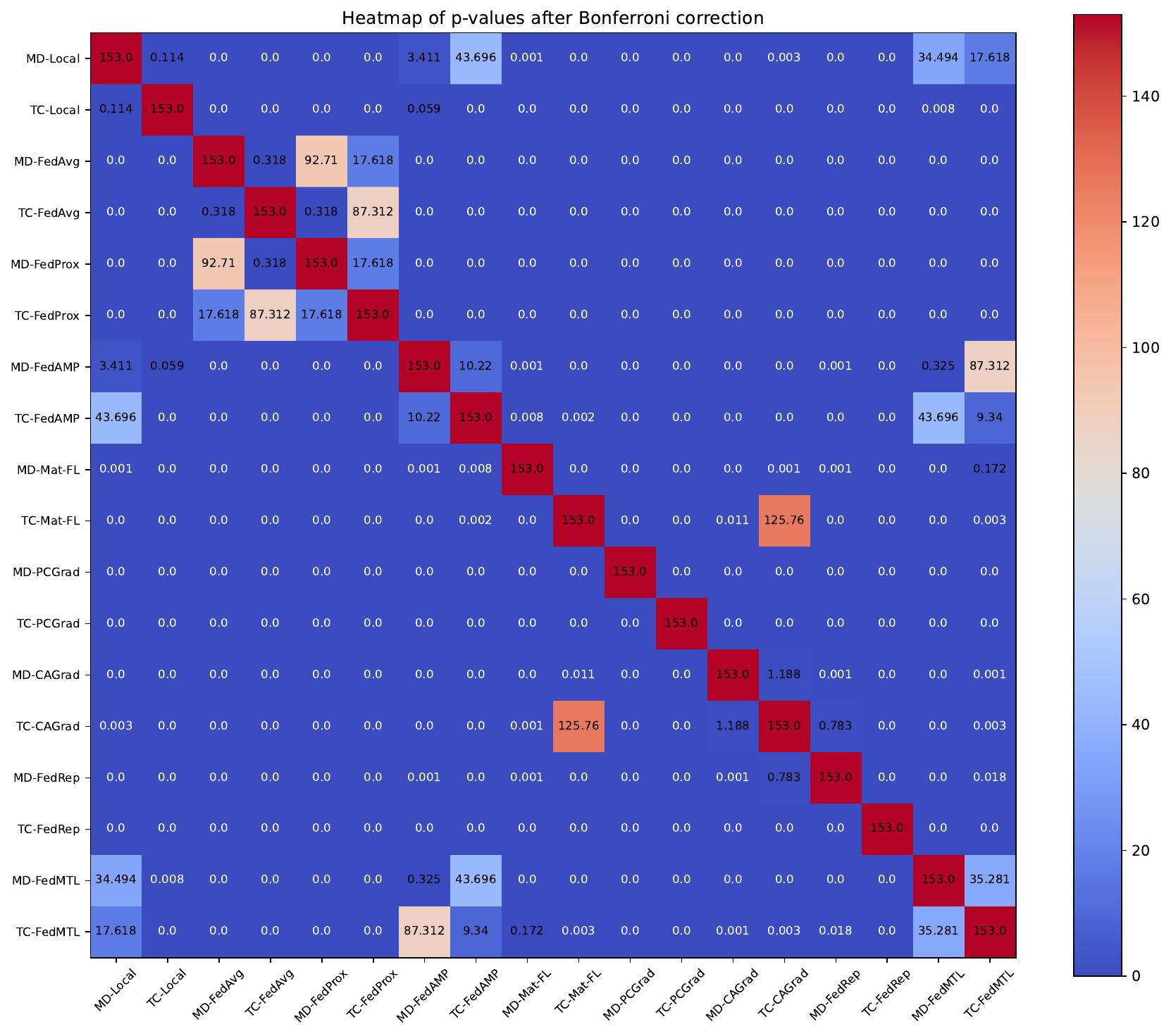}
    \vspace{-10pt}
    \caption{Case Study: Adjusted $p$-value heatmap for pairwise comparisons of baselines in IID-1 SDMT scenario from~\cref{tab: Comp Exp IID-1 and NIID-3}. }
    \begin{flushleft}
        Heatmap displays adjusted $p$-values from pairwise comparisons, utilizing Wilcoxon signed-rank test with Bonferroni correction for multiple comparisons. Each cell corresponds to $p$-value from comparing the row and column baselines. $p$-values below 0.05, marked in white, signify a statistically significant performance difference between the pair of baselines. Please zoom in for details.
    \end{flushleft}
    \label{fig: eval heatmap}
\end{figure}

\subsection{Case Study}
Taking advantage of the unique characteristics of both multi-task learning and federated learning, we perform a case study using a variety of evaluation methods to assess baseline performances. These evaluation methods serve as a robust means for comparing the effectiveness of various algorithms and techniques. The case study complements the comparative experiments as an essential supplement (see~\cref{sec:comp. exp Eval}).

\subsubsection{Additional Evaluation Criteria in the Case Study}
We have selected the IID-1 SDMT scenario in~\cref{tab: Comp Exp IID-1 and NIID-3} as the primary focus of our case study. To meet the practical requirements of FMTL, we have structured our case study as follows:

\noindent\textbf{Performance Improvement Over Time}: Initially, we generate a curve that represents the evolution of the average per-task performance improvement, as defined by Eq. \eqref{eq:avg perform improvement}, as the number of federated learning communication rounds varies. This curve serves as a foundation for discussing the necessary number of rounds to achieve a specific target (see~\cref{fig:eval IID-1 performance improvement}).

\noindent\textbf{Metrics Recording and Analysis}: We record key metrics such as communication overhead, energy consumption, and carbon emissions for each algorithm baseline. We then analyze these metrics in relation to the convergence speed of the algorithms (see~\cref{tab: time comm energy consume}).

\begin{figure}[htbp]
    \centering
    \includegraphics[width=1\linewidth]{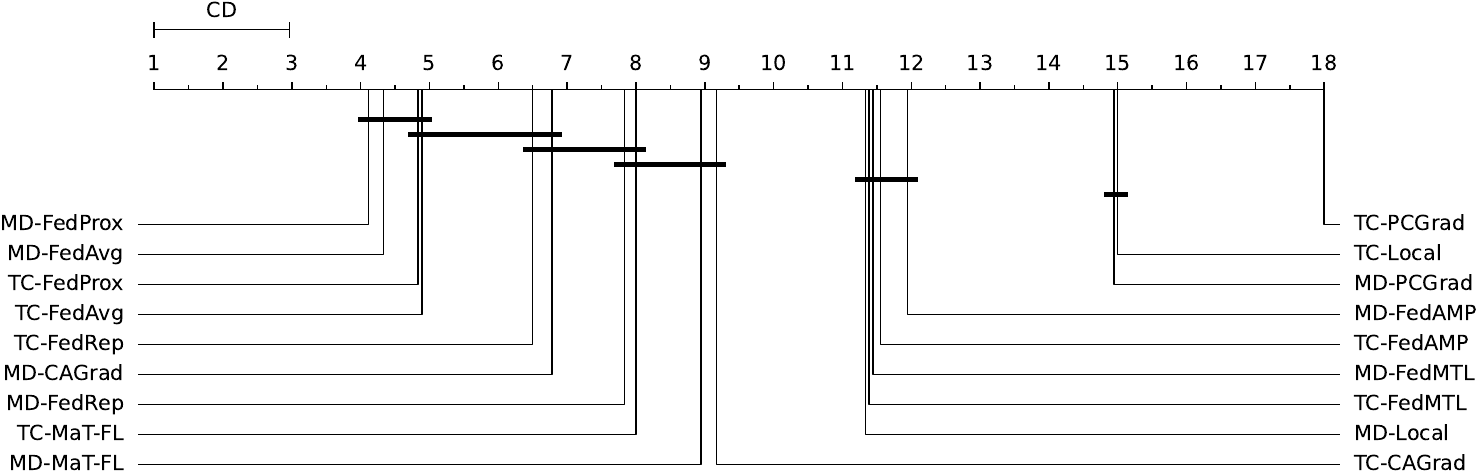}
    \vspace{-10pt}
    \caption{Case Study: the Critical Difference (CD) Diagram for baselines in IID-1 SDMT scenario from Table~\ref{tab: Comp Exp IID-1 and NIID-3}. }
    \begin{flushleft}
    The CD Diagram~\cite{StatComp2006} illustrates the performance-based ranking of various baselines. Sorted from left (best) to right, baselines that are not significantly different according to Nemenyi post-hoc test~\cite{sachs2013angewandte} are connected by horizontal lines. Please zoom in for details.
    \end{flushleft}
    \label{fig: CD}
\end{figure}

\noindent\textbf{Baseline Comparisons}: Acknowledging the unique attributes of different MTL indicators, we first conduct a comparison of baselines using the average per-task performance improvement, a normalized singular metric. We also employ statistical methods, such as the Critical Difference (CD) Diagram~\cite{StatComp2006} with the Nemenyi post-hoc test~\cite{sachs2013angewandte} (see \cref{fig: CD}) for comparing baselines across multiple metrics.

\noindent\textbf{Influence of Pre-Training Strategy and Scalability}: We include experiments where the model is trained from scratch.
We also conduct experiments with varying numbers of clients, ranging from 2 to 8 (see~\cref{fig: case study Scale and train from scratch}).

\begin{table}[htbp]
\setlength{\tabcolsep}{2pt} 
\centering
\caption{Case Study: Comparison of communication costs (GB), energy consumption (kWh), training duration (min), and carbon dioxide emissions (g) for each baseline in the IID-1 Single-Domain Multi-Task (SDMT) scenario from Table~\ref{tab: Comp Exp IID-1 and NIID-3}.}
\vspace{-10pt}
\begin{flushleft}
    \small
    $P$ is equipment running average power (W). $C_r$ and $T_r$ are communication overhead and running time (s) for each round respectively. $R_{30}$, $E_{30}$, $C_{30}$, and $T_{30}$ are respectively the number of communication rounds, energy consumption, communication costs, and time when the average task improvement reaches -30\%, while $R_{10}$, $E_{10}$, $C_{10}$, and $T_{10}$ are for -10\% improvement. $\texttt{CO}_{2}e$ refers to carbon dioxide emissions during training at -10\% improvement. 
    We refer to previous works~\cite{MAS,carbonFootprint2023}, and use the tool~\cite{anthony2020carbontracker} to record power, time, and carbon dioxide emissions on a server equipped with two NVIDIA RTX2080Ti GPUs and two Intel E5-2680V4 CPUs
\end{flushleft}
\resizebox{\linewidth}{!}{
\begin{tabular}{@{}l|c|c|cc|cccc|ccccc|c@{}}
\bottomrule
Algo  & Ar                         & $P$ (w)                          & $C_r$                      & $T_r$ (s)                        & $R_{30}$                        & $E_{30}$                    & $C_{30}$                      & $T_{30}$                  & $R_{10}$                        & $E_{10}$                   & $C_{10}$                      & $T_{10}$                   & \multicolumn{1}{c|}{$\texttt{CO}_{2}e$} & ${\Delta_\textsc{G}\%\uparrow}$                            \\ \hline
                          & MD                         & 738.36                         & 0.00                         & 186.5                         & 11                         & 0.42                         & 0.00                          & 34                         & 23                         & 0.88                         & 0.00                          & 71                          & 476.25                                & 0.00                           \\
\multirow{-2}{*}{Local}   & \cellcolor[HTML]{E7E6E6}TC & \cellcolor[HTML]{E7E6E6}699.72 & \cellcolor[HTML]{E7E6E6}0.00 & \cellcolor[HTML]{E7E6E6}209.3 & \cellcolor[HTML]{E7E6E6}11 & \cellcolor[HTML]{E7E6E6}0.45 & \cellcolor[HTML]{E7E6E6}0.00  & \cellcolor[HTML]{E7E6E6}38 & \cellcolor[HTML]{E7E6E6}19 & \cellcolor[HTML]{E7E6E6}0.77 & \cellcolor[HTML]{E7E6E6}0.00  & \cellcolor[HTML]{E7E6E6}66  & \cellcolor[HTML]{E7E6E6}418.31        & \cellcolor[HTML]{E7E6E6}-1.76  \\ \hline
                          & MD                         & 745.99                         & 1.17                         & 189.0                         & 11                         & 0.43                         & 12.85                         & 35                         & 23                         & 0.90                         & 26.87                         & 72                          & 487.62                                & 13.52                          \\
\multirow{-2}{*}{FedAvg}  & \cellcolor[HTML]{E7E6E6}TC & \cellcolor[HTML]{E7E6E6}705.58 & \cellcolor[HTML]{E7E6E6}0.57 & \cellcolor[HTML]{E7E6E6}210.3 & \cellcolor[HTML]{E7E6E6}15 & \cellcolor[HTML]{E7E6E6}0.62 & \cellcolor[HTML]{E7E6E6}8.62  & \cellcolor[HTML]{E7E6E6}53 & \cellcolor[HTML]{E7E6E6}23 & \cellcolor[HTML]{E7E6E6}0.95 & \cellcolor[HTML]{E7E6E6}13.22 & \cellcolor[HTML]{E7E6E6}81  & \cellcolor[HTML]{E7E6E6}513.06        & \cellcolor[HTML]{E7E6E6}15.15  \\ \hline
                          & MD                         & 720.21                         & 1.17                         & 196.0                         & 11                         & 0.43                         & 12.85                         & 36                         & 23                         & 0.90                         & 26.87                         & 75                          & 488.21                                & 13.55                          \\
\multirow{-2}{*}{FedProx} & \cellcolor[HTML]{E7E6E6}TC & \cellcolor[HTML]{E7E6E6}683.44 & \cellcolor[HTML]{E7E6E6}0.57 & \cellcolor[HTML]{E7E6E6}232.8 & \cellcolor[HTML]{E7E6E6}11 & \cellcolor[HTML]{E7E6E6}0.49 & \cellcolor[HTML]{E7E6E6}6.32  & \cellcolor[HTML]{E7E6E6}43 & \cellcolor[HTML]{E7E6E6}19 & \cellcolor[HTML]{E7E6E6}0.84 & \cellcolor[HTML]{E7E6E6}10.92 & \cellcolor[HTML]{E7E6E6}74  & \cellcolor[HTML]{E7E6E6}454.47        & \cellcolor[HTML]{E7E6E6}16.65  \\ \hline
                          & MD                         & 713.05                         & 1.17                         & 189.9                         & 11                         & 0.41                         & 12.85                         & 35                         & 27                         & 1.02                         & 31.55                         & 85                          & 549.73                                & 0.20                           \\
\multirow{-2}{*}{FedAMP}  & \cellcolor[HTML]{E7E6E6}TC & \cellcolor[HTML]{E7E6E6}710.36 & \cellcolor[HTML]{E7E6E6}0.57 & \cellcolor[HTML]{E7E6E6}215.3 & \cellcolor[HTML]{E7E6E6}7  & \cellcolor[HTML]{E7E6E6}0.30 & \cellcolor[HTML]{E7E6E6}4.02  & \cellcolor[HTML]{E7E6E6}25 & \cellcolor[HTML]{E7E6E6}27 & \cellcolor[HTML]{E7E6E6}1.15 & \cellcolor[HTML]{E7E6E6}15.52 & \cellcolor[HTML]{E7E6E6}97  & \cellcolor[HTML]{E7E6E6}620.80        & \cellcolor[HTML]{E7E6E6}-0.03  \\ \hline
                          & MD                         & 736.19                         & 0.35                         & 190.0                         & 11                         & 0.43                         & 3.84                          & 35                         & 27                         & 1.05                         & 9.43                          & 86                          & 567.89                                & 1.64                           \\
\multirow{-2}{*}{MaT-FL}  & \cellcolor[HTML]{E7E6E6}TC & \cellcolor[HTML]{E7E6E6}699.51 & \cellcolor[HTML]{E7E6E6}0.35 & \cellcolor[HTML]{E7E6E6}212.0 & \cellcolor[HTML]{E7E6E6}7  & \cellcolor[HTML]{E7E6E6}0.29 & \cellcolor[HTML]{E7E6E6}2.45  & \cellcolor[HTML]{E7E6E6}25 & \cellcolor[HTML]{E7E6E6}15 & \cellcolor[HTML]{E7E6E6}0.62 & \cellcolor[HTML]{E7E6E6}5.24  & \cellcolor[HTML]{E7E6E6}53  & \cellcolor[HTML]{E7E6E6}334.49        & \cellcolor[HTML]{E7E6E6}4.93   \\ \hline
                          & MD                         & 736.81                         & 1.17                         & 194.8                         & 11                         & 0.44                         & 12.85                         & 36                         & 27                         & 1.08                         & 31.55                         & 88                          & 582.58                                & -6.61                          \\
\multirow{-2}{*}{PCGrad}  & \cellcolor[HTML]{E7E6E6}TC & \cellcolor[HTML]{E7E6E6}694.33 & \cellcolor[HTML]{E7E6E6}0.57 & \cellcolor[HTML]{E7E6E6}211.5 & \cellcolor[HTML]{E7E6E6}27 & \cellcolor[HTML]{E7E6E6}1.10 & \cellcolor[HTML]{E7E6E6}15.52 & \cellcolor[HTML]{E7E6E6}95 & \cellcolor[HTML]{E7E6E6}55 & \cellcolor[HTML]{E7E6E6}2.24 & \cellcolor[HTML]{E7E6E6}31.61 & \cellcolor[HTML]{E7E6E6}194 & \cellcolor[HTML]{E7E6E6}1214.51       & \cellcolor[HTML]{E7E6E6}-23.81 \\ \hline
                          & MD                         & 736.23                         & 1.17                         & 191.8                         & 11                         & 0.43                         & 12.85                         & 35                         & 23                         & 0.90                         & 26.87                         & 74                          & 488.25                                & 3.25                           \\
\multirow{-2}{*}{CAGrad}  & \cellcolor[HTML]{E7E6E6}TC & \cellcolor[HTML]{E7E6E6}703.17 & \cellcolor[HTML]{E7E6E6}0.57 & \cellcolor[HTML]{E7E6E6}212.0 & \cellcolor[HTML]{E7E6E6}11 & \cellcolor[HTML]{E7E6E6}0.46 & \cellcolor[HTML]{E7E6E6}6.32  & \cellcolor[HTML]{E7E6E6}39 & \cellcolor[HTML]{E7E6E6}23 & \cellcolor[HTML]{E7E6E6}0.95 & \cellcolor[HTML]{E7E6E6}13.22 & \cellcolor[HTML]{E7E6E6}81  & \cellcolor[HTML]{E7E6E6}515.57        & \cellcolor[HTML]{E7E6E6}4.38   \\ \hline
                          & MD                         & 738.90                         & 0.35                         & 188.0                         & 11                         & 0.42                         & 3.84                          & 34                         & 23                         & 0.89                         & 8.04                          & 72                          & 480.43                                & 3.00                           \\
\multirow{-2}{*}{FedRep}  & \cellcolor[HTML]{E7E6E6}TC & \cellcolor[HTML]{E7E6E6}706.55 & \cellcolor[HTML]{E7E6E6}0.35 & \cellcolor[HTML]{E7E6E6}208.8 & \cellcolor[HTML]{E7E6E6}7  & \cellcolor[HTML]{E7E6E6}0.29 & \cellcolor[HTML]{E7E6E6}2.45  & \cellcolor[HTML]{E7E6E6}24 & \cellcolor[HTML]{E7E6E6}19 & \cellcolor[HTML]{E7E6E6}0.78 & \cellcolor[HTML]{E7E6E6}6.64  & \cellcolor[HTML]{E7E6E6}66  & \cellcolor[HTML]{E7E6E6}421.39        & \cellcolor[HTML]{E7E6E6}7.22   \\ \hline
                          & MD                         & 692.65                         & 4.67                         & 202.6                         & 11                         & 0.43                         & 51.41                         & 37                         & 23                         & 0.90                         & 107.50                        & 78                          & 485.35                                & 0.07                           \\
\multirow{-2}{*}{FedMTL}  & \cellcolor[HTML]{E7E6E6}TC & \cellcolor[HTML]{E7E6E6}694.80 & \cellcolor[HTML]{E7E6E6}2.30 & \cellcolor[HTML]{E7E6E6}225.5 & \cellcolor[HTML]{E7E6E6}11 & \cellcolor[HTML]{E7E6E6}0.48 & \cellcolor[HTML]{E7E6E6}25.29 & \cellcolor[HTML]{E7E6E6}41 & \cellcolor[HTML]{E7E6E6}23 & \cellcolor[HTML]{E7E6E6}1.00 & \cellcolor[HTML]{E7E6E6}52.87 & \cellcolor[HTML]{E7E6E6}86  & \cellcolor[HTML]{E7E6E6}541.87        & \cellcolor[HTML]{E7E6E6}-0.58  \\ \toprule
\end{tabular}
}
\label{tab: time comm energy consume}
\end{table}

\begin{table}[htbp]
    \centering
    \caption{Case Study: Influence of client size and scratch training strategy on model performance.}
    \vspace{-10pt}
    \begin{flushleft}
    \small
    SN denotes the scenario, while 2C-8C represent training sets evenly distributed among 2 to 8 clients. ATI refers to Average Task Performance Improvement (${\Delta\%\uparrow}$). The blue shading indicates the target baseline from IID-1 SDMT scenario in~\cref{tab: Comp Exp IID-1 and NIID-3}. Different bar colors and lengths in the upper table denote relative improvements of baselines across scenarios. In the lower table, `*' indicates the use of scratch training strategy, with color brightness representing relative improvement. Refer to \cref{tab: Comp Exp IID-1 and NIID-3} for details.    
    \end{flushleft}     
    \includegraphics[width=1\linewidth]{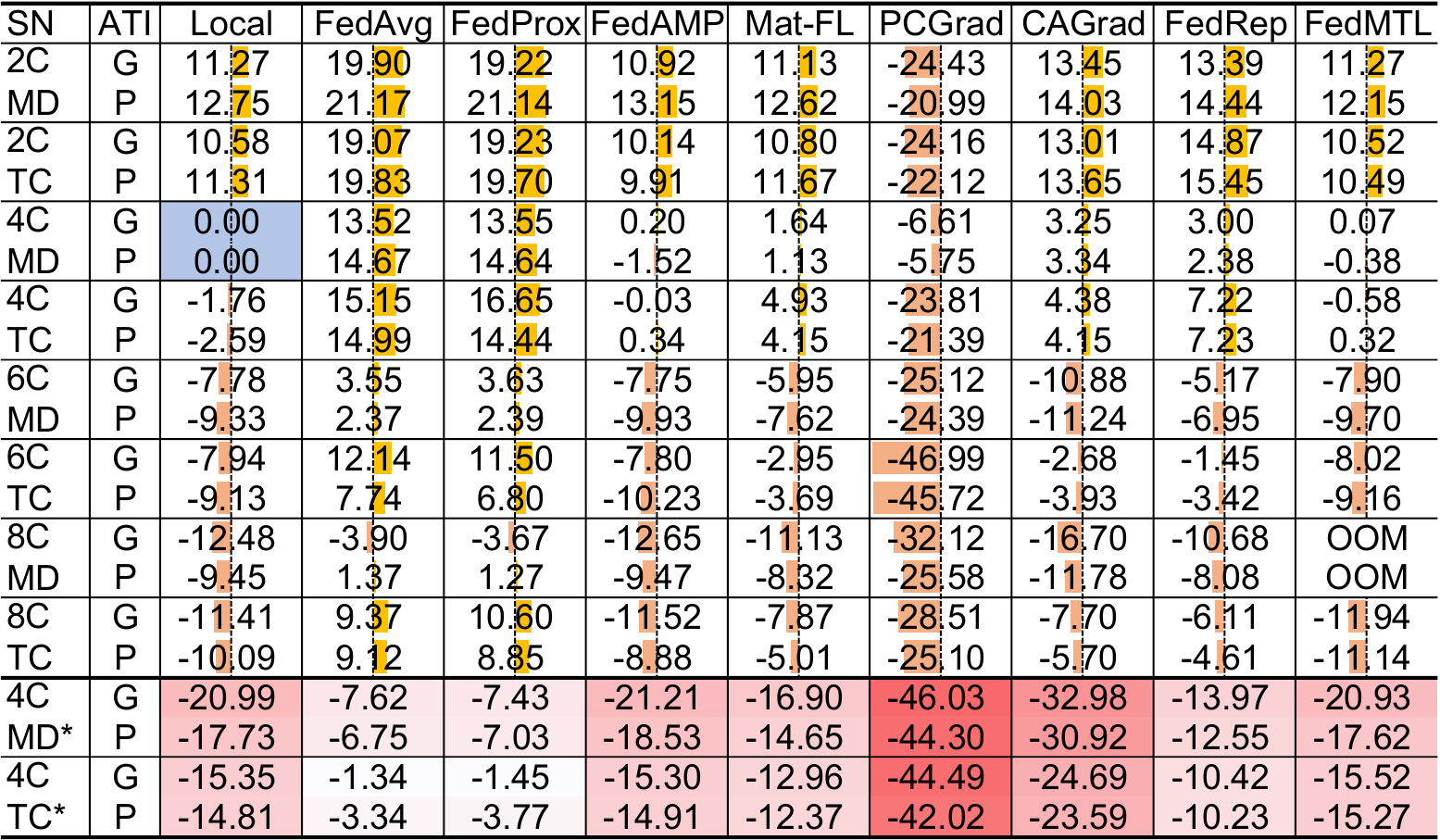}
    \vspace{-10pt}
    \label{fig: case study Scale and train from scratch}
\end{table}

\subsection{Results and Suggestions}
Our experimental findings provide valuable insights and recommendations for future studies and applications in the FMTL.
\subsubsection{General Evaluation}
In accordance with the ``No Free Lunch'' principle, our results from seven comparative experiments demonstrate that no single ``algorithm-model'' baseline consistently outperforms others across all experiments. In fact, the PCGard algorithm, designed to resolve gradient conflict issues, generally underperforms. We advise selecting a baseline method based on the specific requirements of the scenario.

Different types of tasks have distinct indicators, and the relative ratios of their standard deviations also vary considerably. This suggests that the difficulty of learning varies across tasks. Although we treated each task equally in our work, as per~\cref{eq:avg perform improvement}, we posit that in real-world FMTL scenarios, the value of labels for different tasks can vary. Consequently, data holders should select high-value labels based on their actual requirements.

\subsubsection{Data Level Analysis}
In the IID-1 scenario (\cref{tab: Comp Exp IID-1 and NIID-3}, \cref{fig:eval IID-1 performance improvement}, \cref{fig: CD}), FedAvg and FedProx significantly outperform others.

Comparing the Local baseline of IID-1 in~\cref{tab: Comp Exp IID-1 and NIID-3} and NIID-2 in~\cref{tab: Comp Exp NIID-2 SDST}, we observe that in a scenario where the number of labels for each task is balanced: multiple single-task learning models (each model only learns one task) can achieve better performance than a multi-task learning model.
Furthermore, in three sets of pathological partition scenarios (NIID-2, NIID-5 in~\cref{tab: Comp Exp NIID-5 UBSDST}, NIID-7 in~\cref{tab: Comp Exp NIID-7 UBCDST}) where each client has only one task, the combination of FedAvg, FedProx and CAGrad with the SD (single-task MD) has extremely poor performance. FedAMP and FedMTL can achieve a slight G-FL improvement compared to the Local algorithm. The TC architecture suffers less performance loss than the SD architecture. 

In the NIID-3 from~\cref{tab: Comp Exp IID-1 and NIID-3} mixed task number scenario, the combination of FedAvg algorithms with the TC architecture significantly improves the P-FL.   

Compared with IID-1 in~\cref{tab: Comp Exp IID-1 and NIID-3}, the average task improvement indicator of NIID-4 in~\cref{tab: Comp Exp NIID-4 UBSDMT} for algorithms other than FedAvg and FedProx all decreased by 1 to 2 percentage points, suggesting that these two algorithms are relatively robust.

Two sets of cross-domain tasks (NIID-6~\cref{tab: Comp Exp NIID-6 UBCDMT} and NIID-7~\cref{tab: Comp Exp NIID-7 UBCDST}) are jointly trained with the larger PASCAL dataset. Compared with the original IID-1 scenario in \cref{tab: Comp Exp IID-1 and NIID-3}, all baselines perform significantly worse in most tasks when evaluated using each task’s metrics. This indicates cross-domain tasks pose a challenge for FMTL, thereby necessitating design of additional optimization strategies.

\subsubsection{Model Level Analysis}
\textbf{Model Architecture.} As per \cref{tab:parameters and flops}, compared to the widely studied MD architecture \cite{mtlsurvey,MAS}, the TC architecture \cite{astmt, tsn} trades time and computation for space. The MD architecture learns all task labels simultaneously during training, while the TC learns different task labels sequentially. Hence, the TC architecture utilizes fewer model parameters but significantly increases the computational load and training time (see \cref{tab: time comm energy consume}). It also does not require a balanced number of task labels, making it more flexible. For FMTL scenarios, the computing and communication capabilities of the participants and the characteristics of their datasets are crucial considerations, and the two architectures offer different selection biases. 
\textbf{Backbone Network.} It's noteworthy that using a larger backbone significantly improves model performance across all comparative experiments. Provided that device computing capability and inter-client communication bandwidth are sufficient, we recommend using a larger backbone network in FMTL scenarios to handle complex tasks.

\subsubsection{Optimization Algorithm Level Analysis}
As can be seen from NIID-6~\cref{tab: Comp Exp NIID-6 UBCDMT}, the \textbf{parameter decoupling strategy} can aid the MD architecture in performing FL in heterogeneous model scenarios. Simultaneously, it can resist optimization direction conflicts from the MTL process and reduce model performance losses. Also, as observed from the~\cref{tab: time comm energy consume} experiment, it can significantly reduce communication expenses when the same accuracy is achieved.

\subsubsection{Case Study}
As illustrated in \cref{fig: case study Scale and train from scratch}, \textbf{Pre-training Strategy.} Initiating training with a pre-trained model, rather than starting from scratch, can markedly improve the model’s training efficacy. This strategy can significantly decrease communication overhead, training time, and energy consumption.
\textbf{Scale Impact.} When client data diminishes and number of clients escalates, the performance of all algorithms, excluding FedAvg and FedProx, declines significantly. These two algorithms exhibit notable performance improvements compared to Local baseline (especially when combined with TC). This suggests that in IID scenarios, FedAvg and FedProx are highly effective, and TC can also be a primary consideration. 

FMTL encompasses numerous model performance evaluation metrics, with each task having separate indicators. In addition to the average task improvement as in \cref{eq:avg perform improvement}, we introduce a statistical method in \cref{fig: CD} to evaluate the ranking of different baselines across multiple indicators, thereby gaining a clear understanding of the strengths, weaknesses, and statistical differences among baselines.

FMTL also needs to consider real-world deployment issues. The average task improvement as a function of communication rounds was obtained in \cref{fig:eval IID-1 performance improvement} and used in \cref{tab: time comm energy consume}. In the latter, we comprehensively assess the energy consumption, carbon emissions, communication volume, and time consumption of all baselines. Different baselines utilize the federated learning system differently, and the resources used to achieve specified goals also vary. In the IID-1 scenario from \cref{tab: Comp Exp IID-1 and NIID-3}, when reaching a -10\% average task improvement, except for the outlier ``PCGrad-TC'', the time, energy consumption, and carbon emissions of all other baselines are relatively close. The parameter decoupling strategy significantly reduces the communication cost. Future research could incorporate communication expenditure, energy consumption, and time as optimization objectives in real-life FMTL deployment. 
Due to paper length constraints, we have not listed all conclusions from other experiments. These will be discussed in detail in subsequent work.

\section{Conclusion}
The emerging paradigm of federated multi-task learning (FMTL) enables data owners to collaboratively train cross-domain multi-task learning (MTL) models without the need for transferring data from its original domain. To facilitate this, we have developed a benchmark, \algoName, which covers a wide range of settings at data, model, and optimization algorithm levels. We carried out seven sets of comparative experiments to encompass a broad spectrum of data partitioning scenarios. To cater to the practical implementation needs of federated learning (FL) scenarios and the multi-task evaluation requirements of MTL, we utilized a diverse array of evaluation methodologies in our case studies. Through comprehensive experimentation, we have outlined the strengths and weaknesses of existing baseline methods, providing valuable insights for future method selection.

\begin{acks}
This paper is supported by NSFC (No. 62176155), Shanghai Municipal Science and Technology Major Project (2021SHZDZX0102).
\end{acks}

\clearpage
\bibliographystyle{ACM-Reference-Format}
\bibliography{main}

\end{document}